\documentclass[10pt,twocolumn,letterpaper]{article}

\usepackage{cvpr}              

\usepackage{graphicx}
\usepackage{amsmath}
\usepackage{amssymb}
\usepackage{booktabs}
\usepackage{soul}
\usepackage{amssymb}
\usepackage{lipsum}
\usepackage{algorithm}
\usepackage{algpseudocode}
\usepackage{xcolor, colortbl}
\usepackage{mathtools}
\usepackage{tabularx}
\usepackage{multirow}
\usepackage{enumitem}
\usepackage{bbm}
\usepackage{wrapfig}
\usepackage{setspace}
\usepackage{pifont}
\usepackage[accsupp]{axessibility}

\usepackage[pagebackref,breaklinks,colorlinks]{hyperref}

\newcommand{\ours}{FUTR\xspace}
\definecolor{grey}{rgb}{0.9, 0.9, 0.9}
\definecolor{skyblue}{rgb}{0.54, 0.66, 0.87}
\newcommand{\ccol}{\cellcolor{grey}}
\newcommand{\ccolg}{\cellcolor{grey}}

\usepackage{xr}
\makeatletter
\newcommand*{\addFileDependency}[1]{
  \typeout{(#1)}
  \@addtofilelist{#1}
  \IfFileExists{#1}{}{\typeout{No file #1.}}
}
\makeatother

\usepackage{amsmath,amsfonts,bm}

\def\eqref#1{equation~\ref{#1}}

\def\1{\bm{1}}

\DeclareMathAlphabet{\mathsfit}{\encodingdefault}{\sfdefault}{m}{sl}
\SetMathAlphabet{\mathsfit}{bold}{\encodingdefault}{\sfdefault}{bx}{n}

\usepackage[capitalize]{cleveref}
\crefname{section}{Sec.}{Secs.}
\Crefname{section}{Section}{Sections}
\Crefname{table}{Table}{Tables}
\crefname{table}{Tab.}{Tabs.}

\def\ie{\emph{i.e.}}
\def\eg{\emph{e.g.}}
\def\etal{et~al.}

\definecolor{brown}{rgb}{0.65, 0.16, 0.16}
\definecolor{purp}{rgb}{0.65, 0.16, 0.65}
\definecolor{orange}{rgb}{1.0, 0.5, 0.0}
\definecolor{blue}{rgb}{0.0, 0.5, 1.0}
\definecolor{green}{rgb}{0.0, 0.8, 0}
\definecolor{lgreen}{rgb}{0.6, 0.8, 0}
\definecolor{red}{rgb}{0.8, 0, 0}
\definecolor{darkblue}{rgb}{0, 0.2, 0.6}

\newcolumntype{C}[1]{>{\centering\let\newline\\\arraybackslash\hspace{0pt}}p{#1}}

\newcommand{\Sec}[1]{Sec.~\ref{sec:#1}}

\def\cvprPaperID{****} 
\def\confName{CVPR}
\def\confYear{2022}

\begin{document}

\title{Future Transformer for Long-term Action Anticipation}

\author{Dayoung Gong$^1$ \quad Joonseok Lee$^1$ \quad Manjin Kim$^1$ \quad Seong Jong Ha$^2$ \quad Minsu Cho$^1$\vspace{0.15cm}\\
\mbox{POSTECH$^1$ \quad \quad \quad NCSOFT$^2$}\\
{\small \url{http://cvlab.postech.ac.kr/research/FUTR}}
}

\maketitle

\begin{abstract}

The task of predicting future actions from a video is crucial for a real-world agent interacting with others. 
When anticipating actions in the distant future, 
we humans typically consider long-term relations over the whole sequence of actions, i.e., not only observed actions in the past but also potential actions in the future. 
In a similar spirit, we propose an end-to-end attention model for action anticipation, dubbed Future Transformer (FUTR), that leverages global attention over all input frames and output tokens to predict a minutes-long sequence of future actions.
Unlike the previous autoregressive models, the proposed method learns to predict the whole sequence of future actions in parallel decoding, enabling more accurate and fast inference for long-term anticipation.
We evaluate our method on two standard benchmarks for long-term action anticipation, Breakfast and 50 Salads, achieving state-of-the-art results.
\end{abstract}

\vspace{-4mm}
\section{Introduction}
Long-term action anticipation from a video is recently emerging as an essential task for advanced intelligent systems. It aims to predict a sequence of actions in the future from a limited observation of past actions in a video.
While there exists a growing body of research on action anticipation, most of the recent work focuses on predicting a single action in a few seconds~\cite{furnari2019would,miech2019leveraging,fernando2021anticipating,girdhar2021anticipative,sener2020temporal,gammulle2019predicting,sener2019zero, roy2021action}. In contrast, long-term action anticipation~\cite{abu2018will,farha2020long,ke2019time,sener2020temporal} aims to predict a minutes-long sequence of multiple actions in the future. 
This task is challenging since it requires learning long-range dependencies between past and future actions.

\begin{figure}[t]
    \centering
    \includegraphics[width=\linewidth]{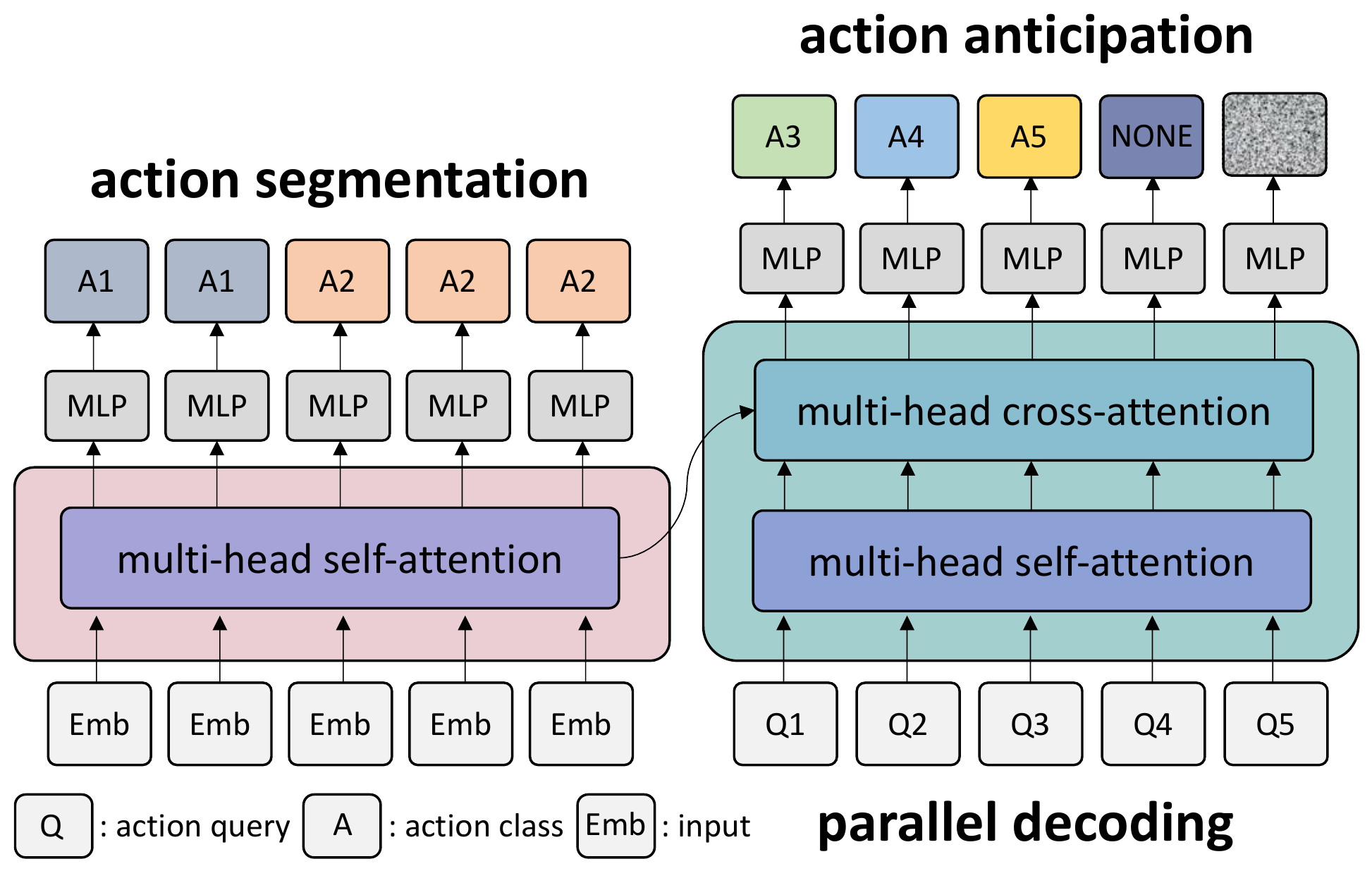}
\caption{\textbf{ Future Transformer (\ours).} 
The proposed method is an end-to-end attention neural network to anticipate actions in parallel decoding, leveraging global interactions between past and future actions for long-term action anticipation.
}
\label{fig:teaser}
\vspace{-2mm}
\end{figure}

Recent long-term anticipation methods~\cite{farha2020long,sener2020temporal} encode observed video frames into condensed vectors and decode them via recurrent neural networks (RNNs) to predict a sequence of future actions in an autoregressive manner.
Despite the impressive performance on the standard benchmarks~\cite{kuehne2014language,stein2013combining}, they have several limitations.
First, the encoder excessively compresses the input frame features so that fine-grained temporal relations between the observed frames are not preserved.
Second, the RNN decoder is limited in modeling long-term dependencies over the input sequence and also in considering global relations between past and future actions.
Third, the sequential prediction of autoregressive decoding may accumulate errors from the precedent results and also increase inference time.
To resolve the limitations, we introduce an end-to-end attention neural network, \textit{Future Transformer (\ours)}, 
for long-term action anticipation. The proposed method effectively captures long-term relations over the whole sequence of actions. 
\ie, not only observed actions in the past but also potential actions in the future.
\ours is an encoder-decoder structure~\cite{vaswani2017attention,carion2020end} as illustrated in Fig.~\ref{fig:teaser}; the encoder learns to capture fine-grained long-range temporal relations between the observed frames from the past, while the decoder learns to capture global relations between upcoming actions in the future along with the observed features from the encoder.
Different from the previous autoregressive models, \ours anticipates a sequence of future actions in parallel decoding, enabling more accurate and faster inference without error accumulations.
Furthermore, we employ an action segmentation loss for input frames to learn distinctive feature representations in the encoder.
We evaluate \ours on the standard benchmarks for long-term action anticipation and achieve new state-of-the-art results on Breakfast and 50 Salads.
The main contribution of our paper is four-fold:
\begin{itemize}
    \item We introduce an end-to-end attention neural network, dubbed \ours, which effectively leverages fine-grained features and global interactions for long-term action anticipation.
    \item We propose to predict a sequence of actions in parallel decoding, enabling accurate and fast inference.
    \item We develop an integrated model that learns distinctive feature representation by segmenting actions in the encoder and anticipating actions in the decoder.
    \item The proposed method sets a new state of the arts on standard benchmarks for long-term action anticipation, Breakfast and 50 Salads.
\end{itemize}

\section{Related Work}

\noindent \textbf{Action anticipation.}
Action anticipation aims to predict future actions given a limited observation of a video. 
With the emergence of the large-scale dataset~\cite{Damen2018EPICKITCHENS,damen2020rescaling}, many methods have been proposed to solve next action anticipation, predicting a single future action within a few seconds~\cite{furnari2019would,miech2019leveraging,fernando2021anticipating,girdhar2021anticipative,sener2020temporal,gammulle2019predicting,sener2019zero, roy2021action}.
Long-term action anticipation has been recently proposed to predict a sequence of future actions in the distant future from a long-range video~\cite{abu2018will,abu2019uncertainty,ke2019time,farha2020long,sener2020temporal}. 
Farha~\etal~\cite{abu2018will} first introduce the long-term action anticipation task and propose two models, RNN and CNN, to tackle the task.
Farha and Gall~\cite{abu2019uncertainty} introduce a GRU network to model the uncertainty of future activities in an autoregressive way. They predict multiple possible sequences of future actions at test time.
Ke~\etal~\cite{ke2019time} introduce a model that predicts an action in a specific future timestamp without anticipating intermediate actions.
They show that iterative predictions of the intermediate actions cause error accumulations.
Previous methods~\cite{abu2018will,ke2019time,abu2019uncertainty} typically take action labels of observed frames as input, extracting action labels using the action segmentation model~\cite{richard2017weakly}.
In contrast, recent work~\cite{farha2020long,sener2020temporal} uses visual features as input.
Farha~\etal~\cite{farha2020long} propose an end-to-end model of long-term action anticipation, employing the action segmentation model~\cite{farha2019ms} for visual features in training.
They also introduce a GRU model with cycle consistency between past and future actions. 
Sener~\etal~\cite{sener2020temporal} suggest a multi-scale temporal aggregation model 
that aggregates past visual features in condensed vectors and then iteratively predicts future actions using the LSTM network.
The recent work~\cite{farha2020long,sener2020temporal} commonly utilizes RNNs with compressed representation of past frames.
In contrast, we propose an end-to-end attention model that anticipates all future actions in parallel using fine-grained visual features of past frames.

\vspace{1mm}
\noindent
\textbf{Self-attention mechanisms.}
Self-attention~\cite{vaswani2017attention} was initially introduced for neural machine translation to mitigate the problem of learning long-term dependencies in RNNs and has been widely adopted in a variety of computer vision tasks~\cite{dosovitskiy2020image,kim2021relational,strudel2021segmenter,fan2021multiscale}.
Self-attention is effective in learning global interactions among image pixels or patches in image domains~\cite{Bello_2019_ICCV,dosovitskiy2020image, touvron2021training,wu2021cvt,strudel2021segmenter,liu2021swin,ramachandran2019stand,yuan2021tokens,wang2021pyramid}.
Several methods employ attention mechanisms in video domains to model temporal dynamics in short-term~\cite{kim2021relational,wang2018non,gberta_2021_ICML,arnab2021vivit,zhang2021vidtr,fan2021multiscale,patrick2021keeping} and long-term videos~\cite{nawhal2021activity,girdhar2021anticipative,zhu2020actbert,luo2020univl,li2020hero}.
Related to action anticipation, 
Girdhar and Grauman~\cite{girdhar2021anticipative} recently introduce the anticipative video transformer (AVT) that uses a self-attention decoder to predict the next action.
Unlike AVT, which requires autoregressive predictions for long-term anticipation, our encoder-decoder model effectively predicts a minutes-long sequence of future actions in parallel.

\vspace{1mm}
\noindent
\textbf{Parallel decoding.}
The transformer~\cite{vaswani2017attention} is designed to predict outputs sequentially, \ie, autoregressive decoding. Due to the inference cost, which increases with the length of the output sequence, recent methods in natural language processing~\cite{gu2018non,stern2018blockwise} replace autoregressive decoding with parallel decoding. The  transformer models with parallel decoding have also been used for computer vision tasks such as object detection~\cite{carion2020end}, camera calibration~\cite{lee2021ctrl}, and dense video captioning~\cite{wang2021end}.
We adopt it for long-term action anticipation, predicting a sequence of future actions simultaneously. In long-term action anticipation, parallel decoding not only enables faster inference but also captures bi-directional relations among future actions.


\section{Problem Setup}
\label{sec:3}

\begin{figure}[t]
    \centering
    \includegraphics[width=0.98\linewidth]{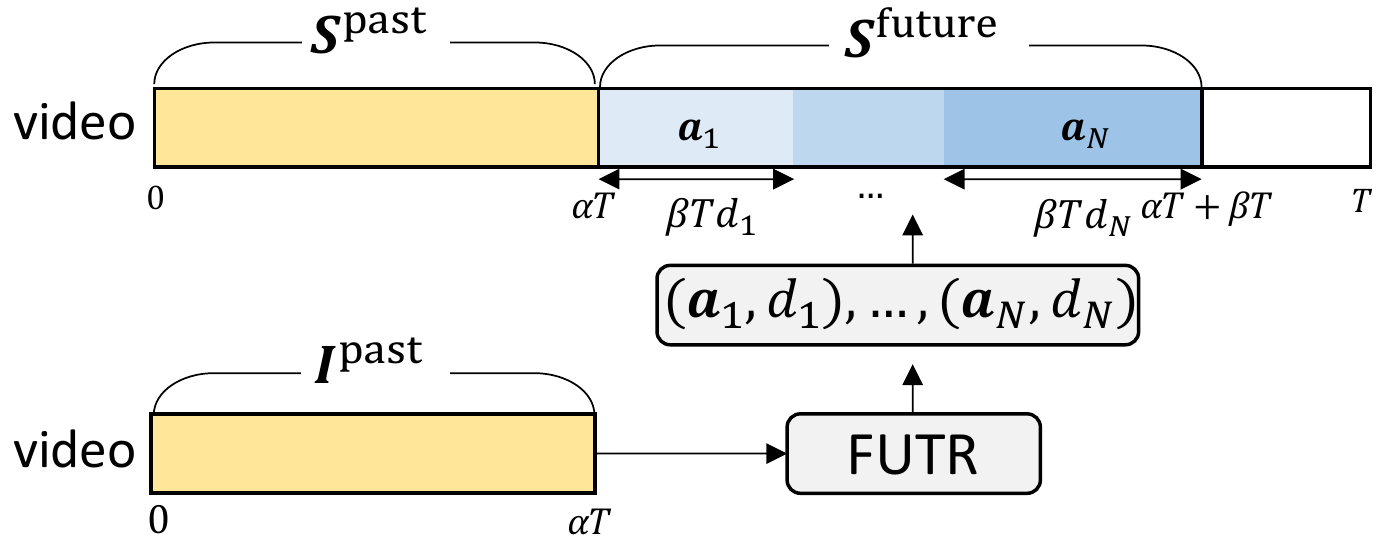}
\caption{\textbf{Long-term action anticipation}.
The problem of long-term action anticipation aims to predict action labels of $\beta T$ future frames observing $\alpha T$ frames from a video, where $\alpha$ and $\beta$ indicate the observation and prediction ratio of the full video frames $T$, respectively. 
\ours anticipates action labels and durations of the $N$ action segments, where predicted action labels and duration are decoded into frame-level action labels for evaluation.}
\label{fig:task}
\vspace{-2mm}
\end{figure}

The problem of long-term action anticipation is to predict a sequence of actions for future video frames from a given observable part of a video.
Figure~\ref{fig:task} illustrates the problem setup.
For a video with $T$ frames, the first $\alpha T$ frames are observed and a sequence of actions for the next $\beta T$ frames is anticipated; $\alpha \in [0,1]$ is an observation ratio of the video while $\beta \in [0, 1-\alpha]$ is a prediction ratio.
The anticipation thus takes the observable frames $\bm{I}^{\mathrm{past}} = [\bm{I}_1, \dots ,\bm{I}_{\alpha T}]^\top \in \mathbb{R}^{\alpha T \times H \times W \times 3}$ as input  
and predicts a sequence of frame-wise action class labels for the next $\beta T$ frames, $\bm{S}^{\mathrm{future}} = [\bm{s}_{\alpha T + 1}, \dots, \bm{s}_{\alpha T+\beta T}]^\top \in \mathbb{R}^{\beta T \times K}$ where $K$ is the number of target actions. 
Following the previous work~\cite{abu2018will, ke2019time, farha2020long, sener2020temporal, abu2019uncertainty}, we represent  $\bm{S}^{\mathrm{future}}$ as a sequence of action segments, each of which consists of an action and its duration, and 
predict a sequence of action class labels $\bm{A}=[\bm{a}_1, \dots, \bm{a}_N]^\top \in \mathbb{R}^{N \times K}$ and their durations $\bm{d}=[d_1, \dots, d_N] \in \mathbb{R}^{N}$ where $\sum^{N}_{j=0} d_j = 1$. 

For evaluation, the sequence of action segments is translated to that of frame-wise actions; the action label $\bm{s}_{\alpha T+t}$ at time $\alpha T+t$ and that of  $i^{\textrm{th}}$ segment are related by 
\vspace{-2mm}
\begin{align}
    \bm{s}_{\alpha T+t} = \bm{a}_i, \quad \beta T \sum^{i-1}_{j=0} d_j < t \leq \beta T \sum^{i}_{j=0} d_j,
    \label{eq:anticipation}
\end{align}
where $d_0=0$.

In addition, the ground-truth action labels for the past frames are denoted by $\bm{S}^{\mathrm{past}} = [\bm{s}_1, \dots ,\bm{s}_{\alpha T}]^\top \in \mathbb{R}^{\alpha T \times K}$, which are used for action segmentation loss in our work. 

\section{Future Transformer~(\ours)}
\begin{figure*}[t!]
    \centering
    \includegraphics[width=\linewidth]{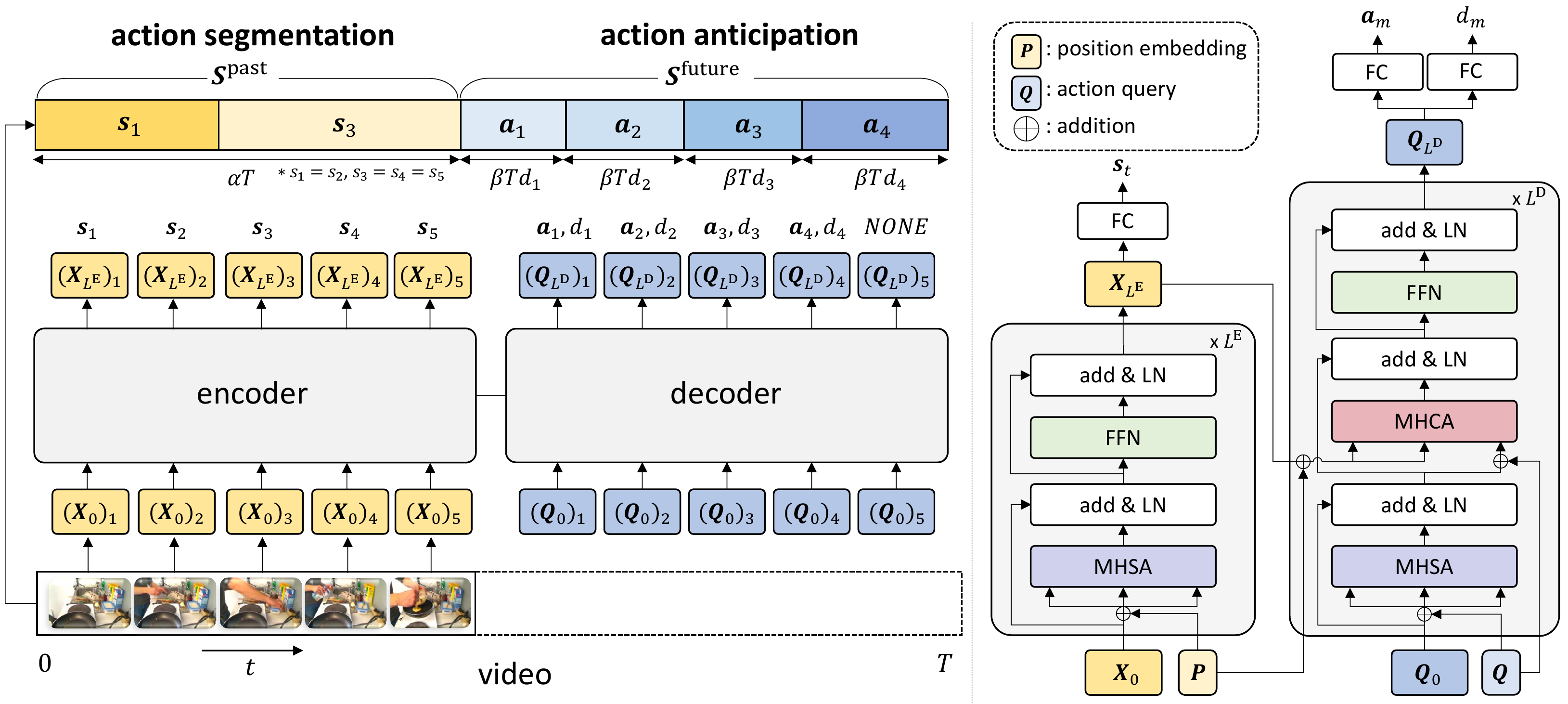}
\caption{\textbf{Overall architecture of \ours.} The proposed method is composed of an encoder and a decoder; each classifies action labels of past frames~(action segmentation) and anticipates future action labels and corresponding durations~(action anticipation), respectively.
The encoder learns distinctive feature representation from past actions via self-attention, and the decoder learns long-term relations between past and future actions via self-attention and cross-attention.
For simplicity, we set the number of past frames $\alpha T$ as 5 and the number of object queries $M$ as 5 in this figure. Note that $(X_l)_i$ and $(Q_l)_i$ indicate $i^\mathrm{th}$ index of $X_l$ and $Q_l$, respectively. }
\label{fig:pipeline}
\vspace{-1mm}
\end{figure*}

In this section, we introduce a fully attention-based network, dubbed \ours, for long-term action anticipation.
The overall architecture consists of a transformer encoder and a decoder, as depicted in Fig.~\ref{fig:pipeline}.
Section~\ref{sec:4.1} explains the encoder, which segments action labels from the fine-grained visual features of past frames, Section~\ref{sec:4.2} describes the decoder, which predicts action labels and durations of future frames in parallel decoding, and then Section~\ref{sec:4.3} presents the training objective of the proposed method.

\subsection{Encoder}
\label{sec:4.1}
The encoder takes visual features as input and segments actions of past frames, learning distinctive feature representations via self-attention.
\\
\noindent
\textbf{Input embedding.}
As input to the encoder, we use visual features extracted from the input frames $\bm{I}^{\mathrm{past}}$, which are denoted by $\bm{F}^{\mathrm{past}} \in \mathbb{R}^{\alpha T \times C}$~\cite{farha2020long,sener2020temporal}.
We sample frames with a temporal stride of $\tau$, establishing $\bm{E} \in \mathbb{R}^{T^\mathrm{O} \times C}$ where $T^\mathrm{O}= \lfloor \frac{\alpha T}{\tau} \rfloor$ is the number of sampled frames.
The sampled frame features are fed to linear layer $\bm{W}^\mathrm{F} \in \mathbb{R}^{C \times D}$ followed by ReLU activation function to $\bm{E}$, creating input tokens $\bm{X}_0 \in \mathbb{R}^{T^\mathrm{O} \times D}$:
\begin{align}
    \bm{X}_0 = \mathrm{ReLU}(\bm{E} \bm{W}^\mathrm{F}).
    \label{eq:1}
\end{align}

\noindent
\textbf{Attention.}
The encoder consists of the $L^\mathrm{E}$ number of encoder layers.
Each encoder layer is composed of a multi-head self-attention (MHSA), layer normalization~(LN) and feed-forward networks~(FFN) with residual connection.
We define a multi-head attention~(MHA) based on the scaled dot-product attention~\cite{vaswani2017attention} with input variables $\bm{X}$ and $\bm{Y}$: 
\begin{align}
    \mathrm{MHA}(\bm{X},\bm{Y})&=[\bm{Z}_1,..,\bm{Z}_h]\bm{W}^\mathrm{O},
    \label{eq:2}
    \\
    \bm{Z}_i&=\mathrm{ATTN}_i(\bm{X},\bm{Y}),
    \label{eq:3}
    \\
    \mathrm{ATTN}_i(\bm{X},\bm{Y})&=\sigma(\frac{(\bm{X}\bm{W}^\mathrm{Q}_i)(\bm{Y}\bm{W}^\mathrm{K}_i)^\top}{\sqrt{D/h}})\bm{Y}\bm{W}^\mathrm{V}_i,
    \label{eq:4}
\end{align}
where $\bm{W}^\mathrm{Q}_i, \bm{W}^\mathrm{K}_i$ and $\bm{W}^\mathrm{V}_i  \in \mathbb{R}^{D \times D/h}$ are query, key, and value projection layer at $i^{\textrm{th}}$ head, respectively, $\bm{W}^\mathrm{O} \in \mathbb{R}^{D \times D}$ is an output projection layer, $h$ is the number of heads, and $\sigma$ indicates a softmax.
MHSA is based on MHA with the two same inputs:
\begin{align}
    \mathrm{MHSA}(\bm{X})&=\mathrm{MHA}(\bm{X},\bm{X}).
    \label{eq:5}
\end{align}
The output token $X_{l+1}$ is obtained from the $l^{\textrm{th}}$ encoder layer:
\begin{align}
    \bm{X}_{l+1} &= \mathrm{LN}(\mathrm{FFN}(\bm{X}_l') + \bm{X}_l'),
    \\
    \bm{X}_l' &= \mathrm{LN}(\mathrm{MHSA}(\bm{X}_l+\bm{P}) + \bm{X}_l),
\end{align}
where an absolute 1-D positional embeddings $\bm{P} \in \mathbb{R}^{T^\mathrm{O} \times D}$ is added to the input of the $l^\mathrm{th}$ layer $\bm{X}_l\in \mathbb{R}^{T^\mathrm{O} \times D}$ for each MHSA layer.

\noindent
\textbf{Action segmentation.}
The final output of the last encoder layer $X_{L^{\mathrm{E}}}$ is utilized to generate action segmentation logits $\hat{\bm{S}}^{\mathrm{past}} \in \mathbb{R}^{T^{\mathrm{O}} \times K}$
by applying a fully-connected~(FC) layer $\bm{W}^{\mathrm{S}}\in\mathbb{R}^{D \times K}$ followed by a softmax:
\begin{align}
    \hat{\bm{S}}^\mathrm{past}&= \sigma(\bm{X}_{L^{\mathrm{E}}}\bm{W}^{\mathrm{S}}).
\end{align}

\subsection{Decoder}
\label{sec:4.2}
The decoder takes learnable tokens as input, referred to as \textit{action queries}, and anticipates future action labels and corresponding durations in parallel, learning long-term relations between past and future actions via self-attention and cross-attention. 
\noindent
\\
\textbf{Action query.}
Action queries are embedded with $M$ learnable tokens $\bm{Q} \in \mathbb{R}^{M \times D}$.
The temporal orders of the queries are fixed to be equivalent to that of the future actions, \ie, the $i^{\textrm{th}}$ query corresponds to the $i^{\textrm{th}}$ future action.
We demonstrate that fixing temporal orders of the queries is effective for long-term action anticipation~(\Sec{5.4}).

\noindent
\textbf{Attention.}
The decoder consists of $L^\mathrm{D}$ number of decoder layers. Each decoder layer is composed of an MHSA, a multi-head cross-attention~(MHCA), LN, and FFN. 
The output query $\bm{Q}_{l+1}$ is obtained from the $l^{\textrm{th}}$ decoder layer:
\vspace{-1mm}
\begin{align}
    {\bm{Q}}_{l+1} &= \mathrm{LN}(\mathrm{FFN}({\bm{Q}}_l'') + {\bm{Q}}_l''),
    \\
    {\bm{Q}}_l'' &= \mathrm{LN}(\mathrm{MHA}(\bm{Q}_l'+\bm{Q},\bm{X}_{L^\mathrm{E}}+\bm{P}) + {\bm{Q}}_l'),
    \\
    {\bm{Q}}_l' &= \mathrm{LN}(\mathrm{MHSA}(\bm{Q}_{l}+\bm{Q})+{\bm{Q}}_l),
    \\
    \bm{Q}_0 &= [0, \dots, 0]^\top \in \mathbb{R}^{M\times D},
\end{align}
where $\bm{X}_{L^\mathrm{E}}$ is the final output of the encoder.
Note that action query $Q$ is added to the input of the $l^{\mathrm{th}}$ layer $Q_l\in\mathbb{R}^{M\times D}$ for each MHSA layer.
We initialize the input of the first decoder layer $Q_0$ with zero vectors.
\begin{table*}[t!]
\vspace{2mm}
\begin{center}
\scalebox{0.95}{
\begin{tabular}{C{1.7cm}C{1.7cm}wl{3cm}C{0.9cm}C{0.9cm}C{0.9cm}C{0.9cm}C{0.9cm}C{0.9cm}C{0.9cm}C{0.9cm}}
\toprule
\multirow{2}*[-0.5ex]{dataset} & \multirow{2}*[-0.5ex]{input type} & \multirow{2}*[-0.5ex]{methods} &\multicolumn{4}{c}{$\beta~(\alpha=0.2)$} & \multicolumn{4}{c}{$\beta~(\alpha=0.3)$}\\
\cmidrule( r){4-7}
\cmidrule( l){8-11}
&&&0.1 & 0.2 & 0.3 & 0.5& 0.1 & 0.2 & 0.3 & 0.5 \\
\cmidrule{1-11}
\multirow{8}*[-0.5ex]{Breakfast} &\multirow{4}*[-0.5ex]{label}& RNN~\cite{abu2018will}  &18.11 & 17.20 & 15.94 & 15.81 &21.64 & 20.02 & 19.73 & 19.21 \\
&&CNN~\cite{abu2018will}        &17.90 & 16.35 & 15.37 & 14.54 &22.44 & 20.12 & 19.69 & 18.76 \\
&&UAAA~(mode)~\cite{abu2019uncertainty} & 16.71 & 15.40 & 14.47 & 14.20 & 20.73 & 18.27 & 18.42 & 16.86 \\
&&Time-Cond.~\cite{ke2019time}  &18.41 & 17.21 & 16.42 & 15.84 &22.75 & 20.44 & 19.64 & 19.75 \\
\cmidrule{3-11}
&\multirow{4}*[-0.5ex]{features}&CNN~\cite{abu2018will} &12.78 & 11.62 & 11.21 & 10.27 & 17.72 & 16.87 & 15.48 & 14.09 \\
&&Temporal Agg.~\cite{sener2020temporal} &24.20 & 21.10 & 20.00 & 18.10 & \underline{30.40} & 26.30 & 23.80 & 21.20 \\
&&Cycle Cons.~\cite{farha2020long}    &\underline{25.88} & \underline{23.42} & \underline{22.42} &\underline{21.54} & 29.66 & \underline{27.37} & \underline{25.58} & \underline{25.20} \\ 
&&\ccol\textbf{\ours~(ours)}     &\ccol \textbf{27.70} & \ccol \textbf{24.55} & \ccol \textbf{22.83} & \ccol \textbf{22.04} &\ccol \textbf{32.27} & \ccol \textbf{29.88} &\ccol \textbf{27.49} &\ccol \textbf{25.87}\\
\cmidrule{1-11}
\multirow{7}*[-0.5ex]{50 Salads}&\multirow{4}*[-0.5ex]{label} & RNN~\cite{abu2018will}  &30.06 & 25.43 & 18.74 & 13.49 & 30.77 & 17.19 & 14.79 & 09.77 \\
&&CNN~\cite{abu2018will}        & 21.24 & 19.03 & 15.98 & 09.87 & 29.14 & 20.14 & 17.46 & 10.86 \\
&&UAAA~(mode)~\cite{abu2019uncertainty} & 24.86 & 22.37 & 19.88 & 12.82 & 29.10 & 20.50 & 15.28 & 12.31\\
&&Time-Cond.~\cite{ke2019time}  &32.51 & 27.61 & 21.26 & \underline{15.99} & \underline{35.12} & \textbf{27.05} & \underline{22.05} & \underline{15.59} \\
\cmidrule{3-11}
&\multirow{3}*[-0.5ex]{features}&Temporal Agg.~\cite{sener2020temporal} &25.50 & 19.90 & 18.20 & 15.10 & 30.60 & 22.50 & 19.10 & 11.20 \\
&&Cycle Cons.~\cite{farha2020long}    &\underline{34.76} & \textbf{28.41} & \underline{21.82} & 15.25 & 34.39 & 23.70 & 18.95 & \textbf{15.89} \\ 
&&\ccol\textbf{\ours~(ours)}     &\ccol \textbf{39.55} & \ccol \underline{27.54} & \ccol \textbf{23.31} & \ccol \textbf{17.77} &\ccol \textbf{35.15} & \ccol \underline{24.86} &\ccol \textbf{24.22} &\ccol 15.26\\
\bottomrule
\end{tabular}
}
\end{center}
\vspace{-4mm}
\caption{\textbf{Comparison with the state of the art.} Our models set a new state of the art on Breakfast, and 50 Salads. The numbers in bold-faced and in underline indicates the highest and the second highest accuracy, respectively.}
\label{tab:sota}
\vspace{-2mm}
\end{table*}

\noindent
\textbf{Action anticipation.} 
The final output of the last decoder layer $Q_{L^{\mathrm{D}}}$ is utilized to generate future actions logits $\hat{\bm{A}} \in \mathbb{R}^{M \times (K+1)}$ by appling a FC layer $\bm{W}^{\mathrm{A}}\in \mathbb{R}^{D \times K+1}$ followed by a softmax and duration vectors $\hat{\bm{d}} \in \mathbb{R}^{M}$ by applying a FC layer $\bm{W}^\mathrm{D}\in \mathbb{R}^{D}$:

\begin{align}
    \hat{\bm{A}}&=\sigma(\bm{Q}_{L^{\mathrm{D}}}\bm{W}^\mathrm{A}),
    \\
    \hat{\bm{d}}&=\bm{Q}_{L^{\mathrm{D}}}\bm{W}^\mathrm{D}.
\end{align}
Note that if none of the future actions are predicted, we let the queries predict a dummy class, `\textit{NONE},' resulting in a total of $K+1$ classes.

\subsection{Training objective}
\label{sec:4.3}
\noindent \textbf{Action segmentation loss.}
We apply action segmentation loss to learn distinctive feature representations of past actions in the encoder as an auxiliary loss.
The action segmentation loss $\mathcal{L}^{\mathrm{seg}}$ is defined with the cross-entropy loss between target actions $\bm{S}^{\mathrm{past}}$ and logits $\hat{\bm{S}}^{\mathrm{past}}$:
\begin{align}
    \mathcal{L}^{\mathrm{seg}} &= -\sum^{T^\mathrm{O}}_{i=1}\sum^{K}_{j=1} \bm{S}^{\mathrm{past}}_{i,j}\log\hat{\bm{S}}^{\mathrm{past}}_{i,j}.
\end{align}
\noindent 
\textbf{Action anticipation losses.}
The $M$ number of action queries are matched to the $N$ number of ground-truth actions to apply action anticipation losses.
Action anticipation loss $\mathcal{L}^{\mathrm{action}}$ is defined with the cross-entropy between target actions $\bm{A}$ and logits $\hat{\bm{A}}$, and duration regression loss $\mathcal{L}^{\mathrm{duration}}$ is defined with L2 loss between target durations $\bm{d}$ and the predicted durations $\hat{\bm{d}}$ :
\begin{align}
    \mathcal{L}^{\mathrm{action}} &= - \sum^{M}_{i=1} \sum^{K+1}_{j=1} \mathbbm{1}_{i \leq \delta} \bm{A}_{i,j}\log(\hat{\bm{A}}_{i,j}),
    \\
    \mathcal{L}^{\mathrm{duration}} &= \sum^M_{i=1} \mathbbm{1}_{\mathrm{argmax}(\hat{\bm{A}}_i) \neq \mathrm{\textit{NONE}}} ( \bm{d}_{i} - \hat{\bm{d}}_i )^2,
\end{align}
where $\delta$ is the position of the first query that predicts \textit{NONE} and $\mathbbm{1}_{i \leq \delta}$ is an indicator function that sets to one where the query position $i$ is less than or equal to $\delta$.
$\mathbbm{1}_{\mathrm{argmax}(\hat{\bm{A}}_i) \neq \mathrm{\textit{NONE}}}$ is also an indicator function that sets to one where the predicted action of the $i^{\mathrm{th}}$ query is not \textit{NONE}.
Note that we apply gaussian normalization to the predicted duration to make summation of the whole durations as 1 following the previous work~\cite{abu2018will,farha2020long}.

\noindent
\textbf{Final loss.} The overall training objective $\mathcal{L}^{\mathrm{total}}$ is the sum of action segmentation loss, action anticipation loss, and duration regression loss:
\begin{align}
    \mathcal{L}^{\mathrm{total}} &=\mathcal{L}^{\mathrm{seg}}+ \mathcal{L}^{\mathrm{action}} + \mathcal{L}^{\mathrm{duration}}.
\end{align}


\section{Experiments}
\subsection{Datasets}
\label{sec:5.1}
We evaluate our method on two standard action anticipation benchmarks: the Breakfast dataset and 50 Salads.

The Breakfast~\cite{kuehne2014language} dataset comprises 1,712 videos of 52 different individuals cooking breakfast in 18 different kitchens.
Every video is categorized into one of the 10 activities related to breakfast preparation.
There exist 48 fine-grained action labels which are used to make up the activities.
On average, each video is about 2.3 minutes long and includes approximately 6 actions.
All videos were down-sampled to a resolution of 240$\times$320 pixels with a frame rate of 15 fps.
The dataset comprises 4 splits of training and test set, and we report the average performance over all the splits following the previous work~\cite{abu2018will,sener2020temporal,farha2020long}.

The 50 Salads~\cite{stein2013combining} dataset comprises 50 videos of 25 people preparing a salad. The dataset contains over 4 hours of RGB-D video data, annotated with 17 fine-grained action labels and 3 high-level activities. Since 50 Salads is usually longer than Breakfast, each video contains 20 actions on average. Every video in the dataset has a resolution of 480$\times$640 pixels with a frame rate of 30 fps. The dataset comprises 5 splits of training and test set, and we report the average results over all the splits.

\subsection{Implementation details}
\label{sec:5.2}
\noindent
\textbf{Architecture details.}
Our model consists of two encoder layers and one decoder layer for Breakfast and two encoder layers and two encoder layers for 50 Salads. 
We set the number of object queries $M$ to 8 for Breakfast and 20 for 50 Salads since 50 Salads includes more actions than Breakfast in a video.
The size of hidden dimension $D$ is set to 128 for Breakfast and 512 for 50 Salads.

\noindent
\textbf{Training \& testing.}
We use pre-extracted I3D features\cite{carreira2017quo} as input visual features for both Breakfast and 50 Salads provided by~\cite{farha2019ms}.
We sample the I3D features with a stride $\tau$ of 3 for Breakfast and 6 for 50 Salads.
In training, we set the observation rate $\alpha \in \{0.2, 0.3, 0.5\}$ and fix the prediction rate $\beta$ to 0.5. 
We use AdamW optimizer\cite{loshchilov2017decoupled} with a learning rate of 1e-3.
We train our model for 60 epochs with a batch size of 16, employing a cosine annealing warm-up scheduler\cite{loshchilov2016sgdr} with warm-up stages of 10 epochs.
In inference, we set the observation rate $\alpha \in \{0.2, 0.3\}$ and prediction rate $\beta \in \{0.1, 0.2, 0.3, 0.5\}$ and measure mean over classes~(MoC) accuracy following the long-term action anticipation framework protocol~\cite{abu2018will,ke2019time,sener2020temporal,farha2020long}.

\subsection{Comparison with the state of the art}
\label{sec:5.3}
In Table~\ref{tab:sota}, we compare our methods with the state-of-the-art methods on Breakfast and 50 Salads.
The table is divided into two compartments according to the dataset, and each compartment is divided into two sub-compartments according to the input types;
the first and the second sub-compartment utilize action labels extracted from the action segmentation model~\cite{richard2017weakly} and visual features, respectively.
For Breakfast, CNN~\cite{abu2018will} uses the Fisher vectors~\cite{abu2018will}, and the other models use I3D features~\cite{carreira2017quo} as input.
For 50 Salads, Sener~\etal~\cite{sener2020temporal} use the Fisher vectors and Farha~\etal~\cite{farha2020long} use I3D features.
As a result, our methods achieve the state-of-the-art performance in all experimental settings on Breakfast and 6 out of 8 settings on 50 Salads, respectively, using visual features only.

\subsection{Analysis}
\label{sec:5.4}
We conduct in-depth analyses to validate the efficacy of the proposed method. In the following experiments, we evaluate our method on the Breakfast dataset setting the observation ratio $\alpha$ as 0.3.
Unless otherwise specified, all experimental settings are the same as those in Sec.~\ref{sec:5.2}.
Further experimental details are indicated in Supp.~A.

\begin{table}[t!]
\begin{center}
\scalebox{0.76}{
\begin{tabular}{C{1.5cm}C{0.5cm}C{1cm}C{0.9cm}C{0.9cm}C{0.9cm}C{0.9cm}C{0.9cm}}
        \toprule
        \multirow{2}*[-0.5ex]{method}&\multirow{2}*[-0.5ex]{AR}& \multirow{2}*[-0.5ex]{\shortstack{causal \\ mask}} & \multicolumn{4}{c}{$\beta~(\alpha=0.3)$} & \multirow{2}*[-0.5ex]{\shortstack{time\\(ms)}}\\
        \cmidrule{4-7}
        &&& 0.1 & 0.2 & 0.3 & 0.5 & \\
        \midrule
        \ours-A &\checkmark&\checkmark& 27.10 & 25.41 & 23.28 & 20.51 & 14.68\\
        \ours-M &-&\checkmark& 31.82 & 28.55 & 26.57 &  24.17 & 5.70\\
        \ccolg \ours &\ccolg-&\ccolg-& \ccolg\textbf{32.27} &\ccolg\textbf{29.88} & \ccolg\textbf{27.49} &  \ccolg\textbf{25.87} & \ccolg \textbf{3.91}\\
        \bottomrule
        \end{tabular}
        }
\caption{\textbf{Parallel decoding vs. autoregressive decoding.}  Parallel decoding significantly improves both accuracy and inference speed. \ours-A autoregressively anticipates future actions using predicted action labels with masked self-attention, and \ours-M anticipates future actions using action queries with masked self-attention. 
All the reported inference speeds are measured on a single RTX 3090 GPU, ignoring data loading time. We feed a single video on GPU and take an average on the test set, except the first ten samples as a warm-up stage for stable inference time~\cite{lin2019tsm}.}
\label{tab:parallel}
\end{center}
\vspace{-4mm}
\end{table}

\vspace{1mm}

\noindent
\textbf{Parallel decoding vs. autoregressive decoding.}
To validate the effectiveness of parallel decoding for long-term action anticipation, we compare our method with two \ours variants with different decoding strategies.
The first variant \ours-A autoregressively anticipates future actions similar to transformer~\cite{vaswani2017attention}. \ours-A takes the output action labels from the previous predictions as input and utilizes masked self-attention. Masked self-attention employs a causal mask to MHSA, which prevents attending to future actions.
The second variant \ours-M is equivalent to \ours except for masked self-attention applied to action queries. \ours-M takes the action queries as input and predicts future actions in parallel, but each query only attends to the past queries.

Table~\ref{tab:parallel} summarizes the results.
FUTR-M outperforms FUTR-A by 3.1-4.7\%p, with 2.6$\times$ faster inference time.
These results demonstrate the effectiveness of parallel decoding using action queries in terms of accuracy and efficiency.
As we remove the causal mask from \ours-M, we obtain an additional accuracy improvement of 0.4-1.7\%p.
Compared to \ours-A, \ours achieves higher accuracy by 4.2-5.4\%p, inferring 3.8$\times$ faster.
These results show the effectiveness of parallel decoding of leveraging bi-directional dependencies between action queries leading to a more accurate and faster inference.
\begin{table}[t!]
\begin{center}
\scalebox{0.9}{
\begin{tabular}{C{1.3cm}C{1.3cm}C{1cm}C{1cm}C{1cm}C{1cm}}
        \toprule
        \multirow{2}*[-0.5ex]{encoder}& \multirow{2}*[-0.5ex]{decoder} & \multicolumn{4}{c}{$\beta~ (\alpha=0.3)$}\\
        \cmidrule{3-6}
        && 0.1 & 0.2 & 0.3 & 0.5 \\
        \cmidrule{1-6}
        LSA&LSA& 27.70 & 24.39 & 23.18 &  21.60\\
        GSA&LSA& 30.15 & 27.51 & 25.62 &  23.28\\
        LSA&GSA& 28.37 & 25.08 & 24.03 &  22.28\\
        \ccolg GSA&\ccolg GSA& \ccolg \textbf{32.27} &\ccolg \textbf{29.88} & \ccolg\textbf{27.49} &  \ccolg\textbf{25.87}\\
        \bottomrule
        \end{tabular}
        }
\caption{\textbf{Global self-attention vs. local self-attention.} Using GSA in both the encoder and the decoder improves the performance, indicating that learning long-term dependencies not only between the observed frames but also between the possible future actions is important for long-term action anticipation.}
\label{tab:locality}
\end{center}
\vspace{-2mm}
\end{table}

\vspace{1mm}
\noindent
\textbf{Global self-attention vs. local self-attention.}
We investigate the effect of learning long-term temporal dependencies between past and future actions by comparing global self-attention~(GSA) and local self-attention~(LSA)~\cite{ramachandran2019stand}.
We build a \ours variant that computes LSA in both the encoder and the decoder, and then replaces LSA with GSA one by one.
We set the window sizes of LSA in the encoder and the decoder as 201 and 3, respectively, where only local area within the window size is utilized for MHA.

The results in Table~\ref{tab:locality} validate the efficacy of using GSA in long-term action anticipation.
From the $1^{\textrm{st}}$ and $2^{\textrm{nd}}$ rows, we observe that replacing LSA with GSA in the encoder improves the accuracy by 1.7-3.1\%p. 
The result verifies that learning the global context between past frames is crucial.
As we replace LSA with GSA in the decoder from the $1^{\textrm{st}}$ and $3^{\textrm{rd}}$ rows, we find consistent accuracy improvement by 0.7-0.9\%p.
We find that learning global temporal relations between action queries is also essential for anticipating a sequence of future actions.
Finally, we replace all LSA to GSA in both the encoder and the decoder from the $1^{\textrm{st}}$ and $4^{\textrm{th}}$ rows, which brings significant improvements by 4.3-5.5\%p, achieving the best accuracy.
These results demonstrate the importance of learning long-term relations between the observed actions in the past and potential actions in the future.

\begin{table}[t!]
\begin{center}
\scalebox{0.77}{
\begin{tabular}{C{1.4cm}C{1.65cm}C{1.6cm}C{0.8cm}C{0.8cm}C{0.8cm}C{0.8cm}}
        \toprule
        \multirow{2}*[-0.5ex]{method}&\multirow{2}*[-0.5ex]{GT Assign.}&\multirow{2}*[-0.5ex]{regression} &  \multicolumn{4}{c}{$\beta~(\alpha=0.3)$}\\
        \cmidrule{4-7}
        &&& 0.1 & 0.2 & 0.3 & 0.5 \\
        \midrule
        \ours-S & sequential  & start-end & 29.15 & 25.51 & 24.20 & 21.43\\
        \ours-H & Hungarian &start-end & 25.26 & 23.85 & 22.63 & 21.45\\
        \ccolg\ours & \ccolg sequential & \ccolg duration & \ccolg\textbf{32.27} &\ccolg\textbf{29.88} & \ccolg\textbf{27.49} &  \ccolg\textbf{25.87}\\
        \bottomrule
        \end{tabular}
        }
\caption{\textbf{Output structuring.}
\ours shows the superior performance over \ours-S and \ours-H. We find that sequential order of the ground truth assignment and duration regression is effective for long-term action anticipation.
Start-end regression predicts normalized start and end timestamp for each query instead of the duration.
}
\label{tab:setpred}
\end{center}
\vspace{-6mm}
\end{table}

\noindent
\textbf{Output structuring.}
To obtain the final output, we consider the action queries as an ordered {\it sequence} and train FUTR to predict an action label and its duration from each in the sequence of action queries; in training, the ground truths of label and duration are directly assigned to the outputs of the queries in the sequential order.
To validate the effectiveness of this output structuring strategy, we compare it with two FUTR variants with different output structuring methods.
\ours-S is a variant of our method that is trained to predict a temporal window of starting and ending points, instead of a duration, from each in the query sequence. 
In inference, the predicted start-end windows are merged with priority according to classification logits; the most confident action labels are assigned to overlapping regions of windows. 
\ours-H is a DETR-like variant~\cite{carion2020end} that considers the action queries as an unordered {\it set}, not a sequence, and predicts a start-end window from each in the query set. In training, the ground truths are assigned to the outputs of the queries by the Hungarian matching~\cite{kuhn1955hungarian}. The matching cost function is defined as the sum of negative class probability and a window loss. 
The Hungarian matching loss is defined as the sum of the action anticipation loss and the window loss. 
See Supp.~A for the details.

Table~\ref{tab:setpred} shows the performances of the two variants and ours.  
The comparison between \ours-S and \ours-H shows that the sequential ground-truth assignment is more effective in training than the Hungarian assignment, which implies that the fixed sequence of action queries facilitates to capture temporal dependencies in a more effective manner. 
The comparison between \ours and \ours-H finds that the duration regression is more effective than the start-end regression, achieving a significant accuracy gain.

\vspace{1mm}

\noindent
\textbf{Loss ablations.}
%
\begin{table}[t!]
\begin{center}
\scalebox{0.88}{
\begin{tabular}{C{0.75cm}C{1.05cm}C{1.4cm}C{0.8cm}C{0.8cm}C{0.8cm}C{0.8cm}}
        \toprule
        \multicolumn{3}{c}{loss} & \multicolumn{4}{c}{$\beta~(\alpha=0.3)$} \\ 
        \cmidrule( r){1-3}
        \cmidrule( l){4-7}
        $\mathcal{L}^{\mathrm{seg}}$&$\mathcal{L}^{\mathrm{action}}$&$\mathcal{L}^{\mathrm{duration}}$ & 0.1 & 0.2 & 0.3 & 0.5 \\
        \cmidrule{1-7}
        -&\checkmark&\checkmark& 28.31 & 25.85 & 24.91 & 22.50\\
        \ccolg\checkmark&\ccolg\checkmark&\ccolg\checkmark & \ccolg \textbf{32.27} & \ccolg \textbf{29.88} &  \ccolg\textbf{27.49} & \ccolg \textbf{25.87}\\
        \bottomrule
        \end{tabular}}
\end{center}
\vspace{-2mm}
\caption{\textbf{Loss ablations.} Action segmentation loss significantly improves the performance, indicating that recognizing past frames is crucial for anticipating future actions.}
\vspace{-4mm}
\label{tab:loss}
\end{table}
In Table~\ref{tab:loss}, we evaluate the effectiveness of action segmentation loss. As a result, action segmentation loss significantly improves the performance, indicating that recognizing past frames plays a crucial role in anticipating future actions.
\begin{figure}[t]
    \centering
    \begin{subfigure}[t]{\linewidth}
    \includegraphics[width=\columnwidth]{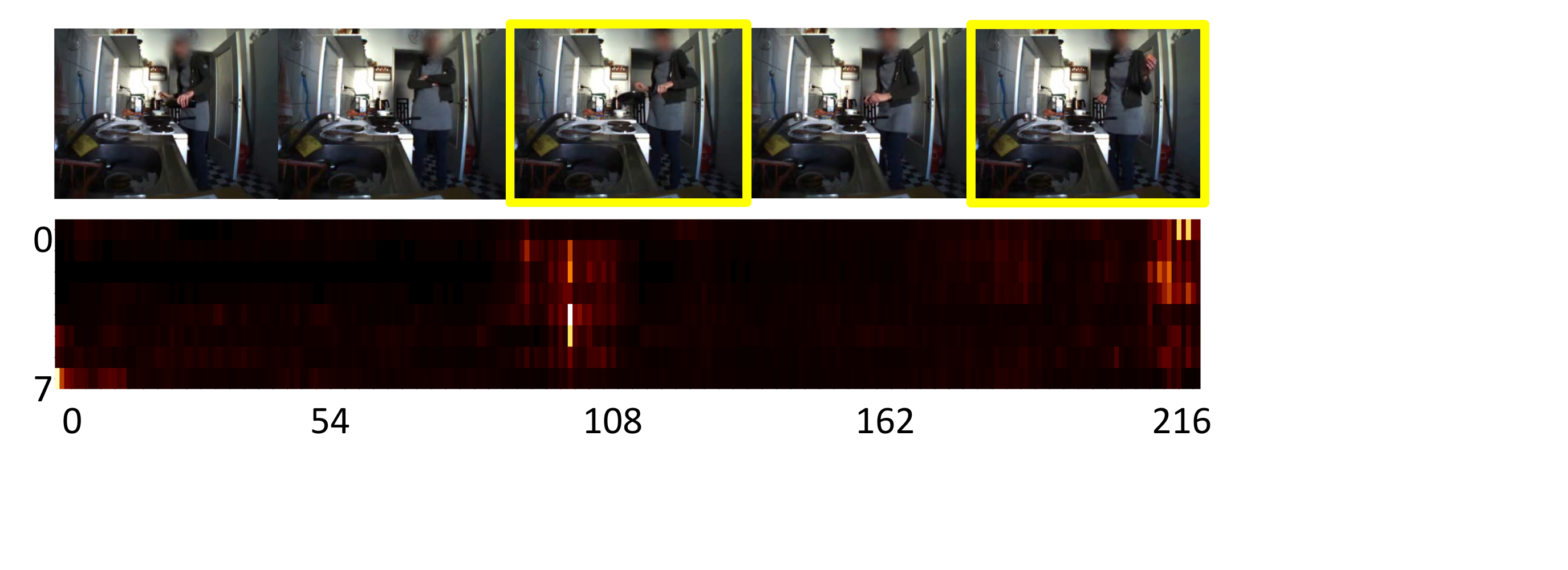}
    \caption{Activity: Fried egg}
    \label{fig:4a}
    \end{subfigure}
    \begin{subfigure}[t]{\linewidth}
    \includegraphics[width=\columnwidth]{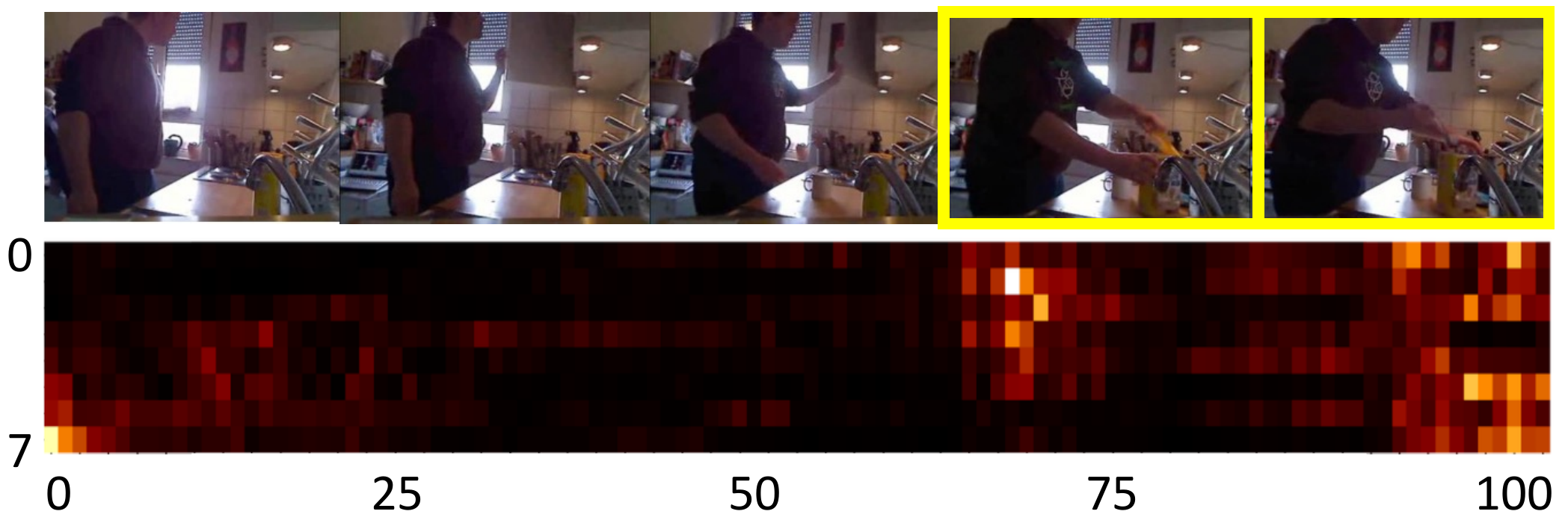}
    \caption{Activity: Milk}
    \label{fig:4b}
    \end{subfigure}
\caption{\textbf{Cross-attention map visualization on Breakfast}. The horizontal and vertical axis indicates the index of the past frames and the action queries, respectively. The brighter color implies a higher attention score. We highlight a video frame with a yellow box where the attention score of the frame is highly activated. RGB frames are uniformly sampled from a video in this figure.}
 \label{fig:4}
\end{figure}
\begin{figure*}[t]
    \centering
    \begin{subfigure}[t]{\linewidth}
    \includegraphics[width=\columnwidth]{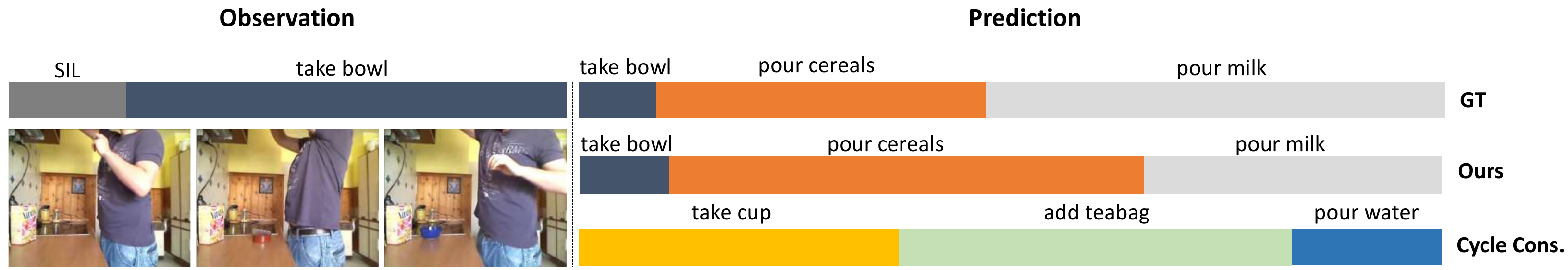}
    \caption{Activity: Cereal}
    \label{fig:5a}
    \end{subfigure}
    \vspace{-1mm}
    \begin{subfigure}[t]{\linewidth}
    \includegraphics[width=\columnwidth]{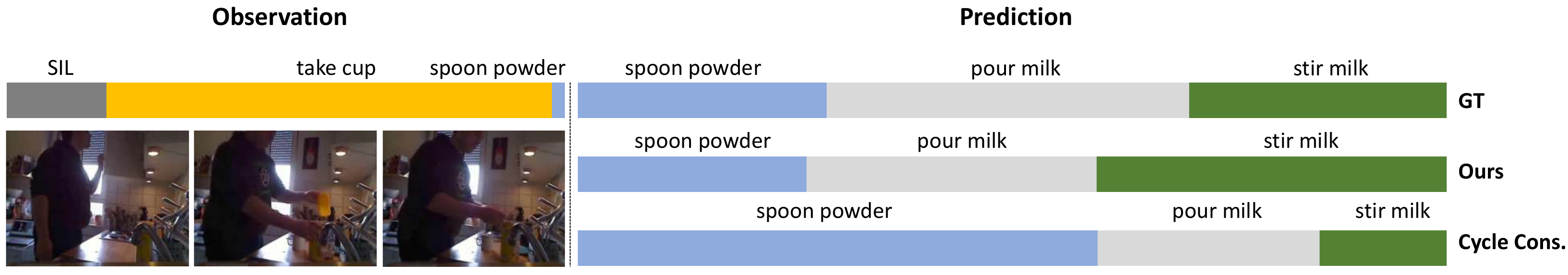}
    \caption{Activity: Milk}
    \label{fig:5b}
    \end{subfigure}
    \vspace{-1mm}
    \begin{subfigure}[t]{\linewidth}
    \includegraphics[width=\columnwidth]{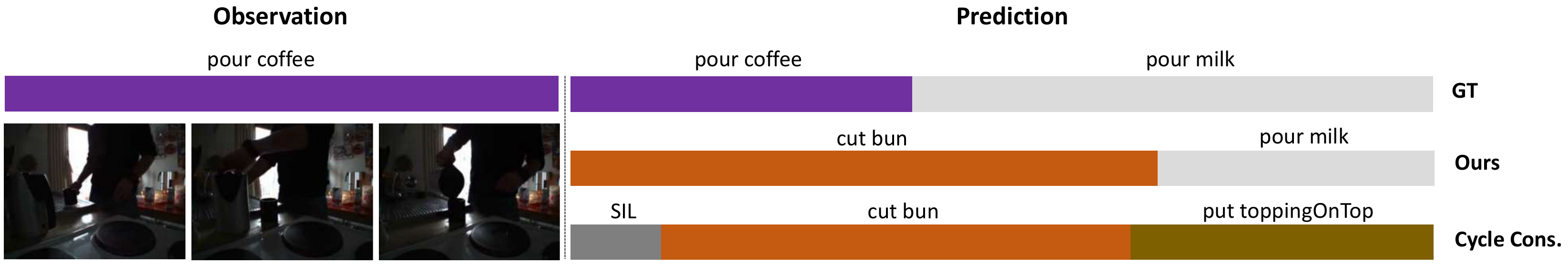}
    \caption{Activity: Coffee}
    \label{fig:5c}
    \end{subfigure}
\caption{\textbf{Qualitative results on Breakfast}. Each subfigure visualizes the ground-truth labels and predicted results of the \ours and the Cycle Cons.~\cite{farha2020long}. We set $ \alpha $ as 0.3 and $ \beta $ as 0.5 in this experiment. We decode action labels and durations as frame-wise action classes. Each color in the color bar indicates an action label written above.}
\vspace{-4mm}
 \label{fig:5}
\end{figure*}
\begin{table}[t!]
\begin{center}
\scalebox{0.9}{
\begin{tabular}{C{1.3cm}C{1.4cm}C{1.4cm}C{1.4cm}C{1.4cm}}
        \toprule
        \multirow{2}*[-0.5ex]{$M$}& \multicolumn{4}{c}{$\beta~(\alpha=0.3)$} \\ 
        \cmidrule( l){2-5}
        & 0.1 & 0.2 & 0.3 & 0.5 \\
        \cmidrule{1-5}
        6 & 29.95 & 26.47 & 25.46 & 23.27 \\
        7 & 30.03 & 27.94& 27.00 & 24.23 \\
        \ccolg 8 & \ccolg \textbf{32.27} & \ccolg \textbf{29.88} &  \ccolg 27.49 & \ccolg \textbf{25.87}\\
        9 & 31.24 & 28.65& 26.87 & 24.95 \\
        10 & 31.32 & 28.86& \textbf{27.74} & 25.01 \\
        \bottomrule
        \end{tabular}}
\end{center}
\vspace{-2mm}
\caption{\textbf{Number of action queries.} We adjust the number of action queries $M$, showing that a sufficient number of action queries shows saturated performance. }
\vspace{-4mm}
\label{tab:query}
\end{table}


\vspace{2mm}

\noindent
\textbf{Number of action queries.}
To analyze the impact of the number of action queries $M$ in \ours, we adjust the value of $M$ from 6 to 10. In Table~\ref{tab:query}, the performance becomes saturated as we gradually increase the number of action queries. By this experiment, we set $M$ to 8 for Breakfast.

See Supp.~C and D for additional analysis and results.

\subsection{Attention map visualization}
We visualize the cross-attention layers in the decoder in Fig.~\ref{fig:4}. The vertical and horizontal axis indicates the index of the action queries and input past frames, respectively. 
We find two interesting results from this experiment.
First, our model learns to attend to visual features in the recent past, showing that the nearest frames provide crucial keys for predicting future actions. 
It is in alignment with the previous work~\cite{ke2019time,sener2020temporal} that reflects the importance of the recent past in designing anticipation models.
\ours also attends to the recent past without any prior knowledge applied to the model.
Second, we find that \ours is trained to attend to important actions not only from the recent past, but also from the entire past frames. In Fig.~\ref{fig:4a}, essential frames with yellow boxes are detected by the queries with the high attention scores, providing contextual clues of the activity,~\eg `holding a pan' and `taking an egg' actions in the `fried egg' activity. Furthermore, action queries anticipating \textit{NONE} class attend to the irrelevant features such as the beginning of the videos. The results show that \ours effectively leverages long-term dependencies using the entire past frames regardless of the position, and also detects key frames of the given activity.
More visualization results are shown in Supp.~E.

\subsection{Qualitative results}
Figure~\ref{fig:5} shows the qualitative results of \ours and Cycle Cons.~\cite{farha2020long}, evaluating on long-term action anticipation. In this experiment, we plot the prediction results based on the a sequence of predicted action label and corresponding duration. Each subfigure consists of observed frames, the ground-truth~(GT) labels, and prediction results from the two models. Observed frames are uniformly sampled from videos.
Figure~\ref{fig:5a} shows the importance of utilizing fine-grained features for action anticipation. 
\ours anticipates `take bowl' action from the observed frames, while Cycle Cons. model anticipates `take cup' action missing fine-grained features, which leads to the error accumulation of the rest of the predictions.
Figure~\ref{fig:5c} validates the robustness of parallel decoding on error accumulations from the previous predictions. Although the two models were wrong in the first anticipation, our model correctly predicts the following action label while Cycle Cons. generates false results during iterative predictions.
The results also validates effectiveness of the proposed methods on various activities.
See Supp.~E for more qualitative results.

\vspace{-2mm}
\section{Conclusion}
We have introduced an end-to-end attention neural network, \ours, which leverages global relations of past and future actions for long-term action anticipation. 
The proposed method utilizes fine-grained visual features as input and anticipates future actions in parallel decoding, enabling accurate and faster inference. 
We have demonstrated the advantages of our method through extensive experiments on two benchmarks, achieving a new state of the art. 
While we have focused on long-term action anticipation in this work, we proposed an integrated model of action segmentation and anticipation in the same framework.
We believe that \ours suggested the direction that enhances comprehension of the actions in long-range videos. 

\vspace{1mm}
\noindent
\textbf{Acknowledgements.} This research was supported by NCSOFT, the IITP grant funded by MSIT~(No.2019-0-01906, AI Graduate School Program - POSTECH), and the Center for Applied Research in Artificial Intelligence (CARAI) grant funded by DAPA and ADD~(UD190031RD). 

{\small
\bibliographystyle{ieee_fullname}
\bibliography{egbib}
}
\clearpage

\twocolumn[{%
\renewcommand\twocolumn[1][]{#1}%
\maketitle
\begin{center}
    \centering
    \vspace{-4mm}
    \counterwithin{figure}{section}
    \captionsetup{type=figure}
    \begin{minipage}{0.27\linewidth}
    \includegraphics[width=\textwidth]{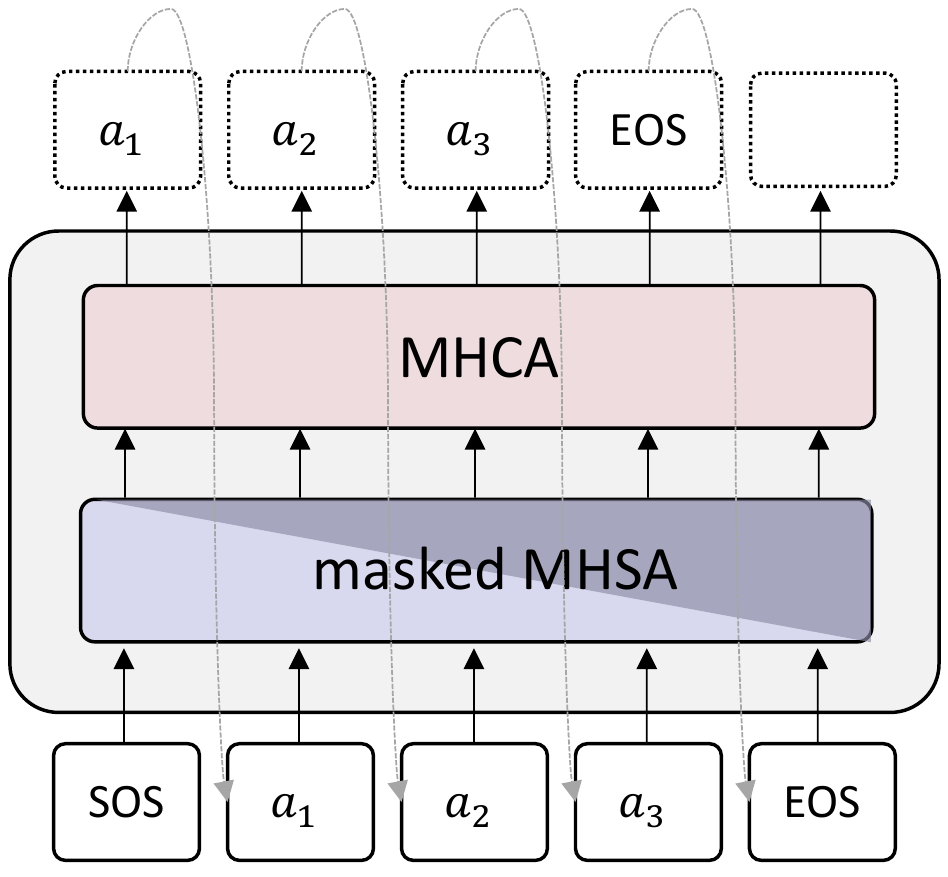}
    \centering\small{(a) FUTR-A}
    \end{minipage}\hfill
    \begin{minipage}{0.27\linewidth}
    \includegraphics[width=\textwidth]{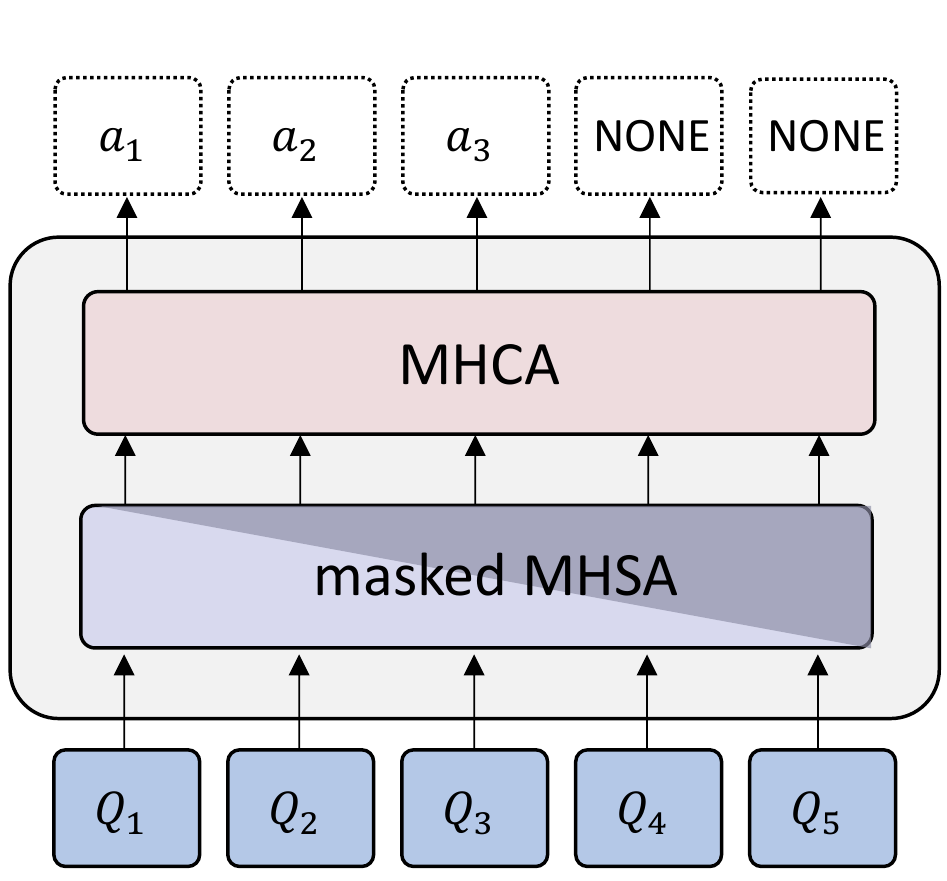}
    \centering\small{(b) FUTR-M}
    \end{minipage}\hfill
    \begin{minipage}{0.27\linewidth}
    \includegraphics[width=\textwidth]{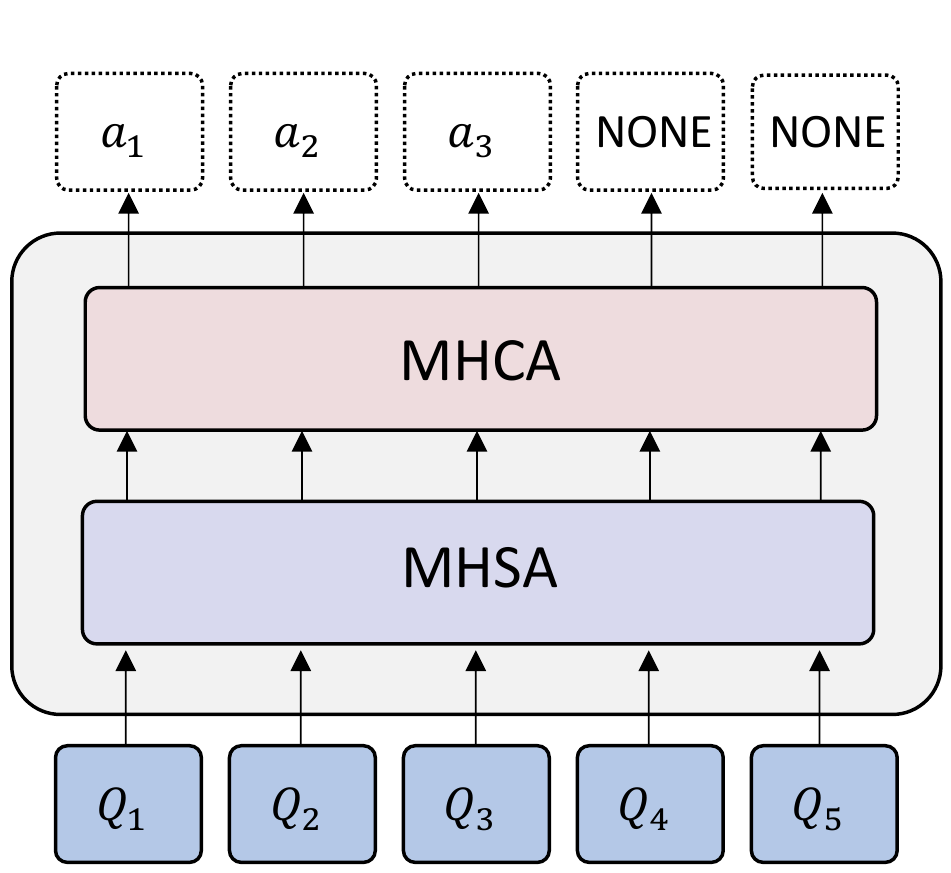}
    \centering\small{(c) FUTR}
    \end{minipage}\hfill
    \begin{minipage}{0.15\linewidth}
    \includegraphics[width=\textwidth]{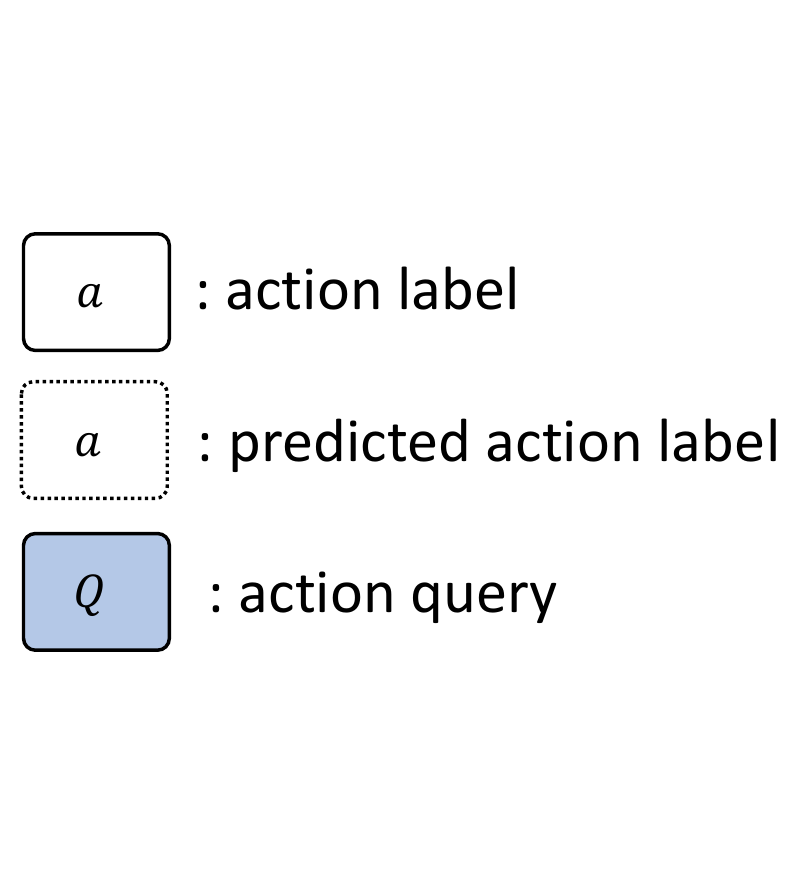}
    \end{minipage}\hfill
    \counterwithin{figure}{section}
    \renewcommand\thefigure{S\arabic{figure}}
    \caption{\textbf{\ours variants with different decoding strategies.}
    (a) \ours-A autoregressively anticipates future actions using the output action labels from the previous predictions as input and utilizes masked self-attention.
    (b) \ours-M is equivalent to \ours except for masked self-attention applied to action queries.
    (c) \ours remove a causal mask in MHSA, where query attends to past and future queries in the sequence. \ours-M and \ours anticipates action and duration for each query simultaneously.
    }
    \label{fig:parallel}
\end{center}%
}]

\section*{\Large{Supplementary Material}}

\section*{A. Experimental Details}
\label{sec:A}
In this section, we provide experimental details of the two experiments in Sec.~\ref{sec:5.4}.
\noindent
\\
\textbf{Parallel decoding vs. autoregressive decoding.}
In Table~\ref{tab:parallel}, we compare our model with two \ours variants, \ours-A and \ours-M.
Two models have the same encoder but different decoders compared to \ours, as illustrated in Fig.~\ref{fig:parallel}.
\ours-A anticipates the next action recurrently using a sequence of the predicted action labels as input in an autoregressive way.
There exist two unique tokens: \textit{SOS} and \textit{EOS} in autoregressive decoding, each of which indicates the start and the end of the sequence, respectively.
The decoder of \ours-A takes \textit{SOS} as the first input and predicts the next action label recursively until the model predicts \textit{EOS}.
\ours-M takes a sequence of action queries as input and predicts action labels and durations in parallel with masked self-attention. Masked self-attention employs a causal mask to MHSA, preventing action queries from attending to future actions.
The core difference between \ours and \ours-M lies in the masked self-attention; action queries of \ours-M only consider uni-directional temporal dependencies between action queries, while that of \ours consider bi-directional temporal relations between the past and the future.  
We validate the effect of parallel decoding by comparing the three models.
\\
\noindent
\textbf{Output structuring.}
In Table~\ref{tab:setpred}, we conduct experiments related to output structuring strategy.
We introduce two variants of \ours, \ours-H and \ours-S. 
\ours-H is a DETR-like variant~\cite{carion2020end}, where the ground truths are assigned to the outputs of the queries by the Hungarian matching~\cite{kuhn1955hungarian}.
Let us denote that $\bm{y}$ is the target set of future actions. 
The ground truth of the $i^{\mathrm{th}}$ index is defined by $\bm{y}_i = \{\bm{c_i}, \bm{t}_i\}$, where $\bm{c_i}$ and $\bm{t}_i$ is the target action label and start-end window, respectively.
Note that $\bm{y}$ is padded with \textit{NONE} class to a size $M$.
We also denote $\hat{\bm{y}}$ is the set of $M$ predictions from the action queries. 
Since the Hungarian matching finds a pair-wise matching between the two set $\bm{y}$ and $\hat{\bm{y}}$ minimizing the matching cost $\mathcal{L}^{\mathrm{match}}$, we find the optimal permutation $\zeta$ from a set of permutation of M queries $Z_M$:
\begin{align}
    \hat{\zeta} &= \mathrm{argmin}_{\zeta \in Z_M}\sum_{i=1}^M\mathcal{L}^{\mathrm{match}}(\bm{y}_i, \hat{\bm{y}}_{\zeta(i)}).
\end{align}
We define matching cost as the sum of negative class probability and a window loss:
\begin{align}
     \mathcal{L}^{\mathrm{match}}(\bm{y}_i, \hat{\bm{y}}_i) &= \mathbbm{1}_{c_i\neq \varnothing}[-\hat{\bm{A}}_{\zeta(i),c_i} +\mathcal{L}^{\mathrm{window}}(\bm{t}_i,\hat{\bm{t}}_{\zeta(i)})],
\end{align}
where $\mathbbm{1}_{c_i\neq\varnothing}$ is an indicator function that sets to one where the gournd-truth action label is not \textit{NONE}.
We define a window loss $\mathcal{L}^{\textrm{window}}$ with L1 distance and temporal IoU loss:
\begin{align}
    \mathcal{L}^\mathrm{window}(\bm{t}_i, \hat{\bm{t}}_{\zeta(i)}) &= \lambda^\mathrm{L_1}\lvert\lvert \bm{t}_i- \hat{\bm{t}}_{\zeta(i)}\rvert\rvert_1 - \lambda^\mathrm{tiou} \frac{\lvert \bm{t}_i \cap \hat{\bm{t}}_{\zeta(i)} \rvert}{\lvert \bm{t}_i \cup \hat{\bm{t}}_{\zeta(i)}\rvert},
\end{align}
where $|.|$ and $\hat{\bm{t}}_{\zeta(i)}$ indicates temporal areas and the predicted start-end window. 
$\lambda^\mathrm{L1}$ and $\lambda^\mathrm{tiou}$ are weighting values of the two losses, which are 5 and 2, respectively.
Note that starting and ending points of the temporal window $\bm{t}_i \in [0,1]^2$ are bounded from 0 to 1.
Finally, we define the Hungarian loss $ \mathcal{L}^\mathrm{Hungarian}$ by 
\begin{multline}
    \mathcal{L}^\mathrm{Hungarian}(\bm{y}, \hat{\bm{y}}) \\
    = \sum_{i=1}^M\sum_{j=1}^{K+1}[-{\bm{A}}_{i,j}log\hat{\bm{A}}_{\hat{\zeta}(i),j}+
        \mathbbm{1}_{c_i\neq \varnothing}\mathcal{L}^{\mathrm{window}}(\bm{t}_i, \hat{\bm{t}}_{\hat{\zeta}(i)})].
\end{multline}
In training \ours-H, we use the sum of the Hungarian loss and the action segmentation loss as our final loss.

\section*{B. Next Action Anticipation}
We conduct an experiment of next action anticipation on EK55~(validation, RGB) following the previous experimental protocols~\cite{girdhar2021anticipative,sener2020temporal,furnari2019would}.
\\
\noindent
\textbf{Dataset.}
The Epic-Kitchens 55 dataset~\cite{Damen2018EPICKITCHENS} is the large-scale dataset in first-person vision. The dataset comprises of 55 hours of recordings of 32 kitchens, including 39,594 action segments annotated with 125 verb, 331 noun, and 2,513 action classes. 
\\
\noindent
\textbf{Implementation details.}
\ours can be applied to next action anticipation by simply setting the number of action query $M$ to 1.
We use two encoder layers and two decoder layers while setting the size of the hidden dimension $D$ to 512.
We do not include action segmentation loss in this experiment due to the
lack of frame-level action annotations. 
Instead, we use additional a fully-connected layer applying to the output of the encoder layers $X_{L^E}$ to predict features of the next frame. Then we apply a feature prediction loss of L2 distance between predicted features and the next frame similar to AVT~\cite{girdhar2021anticipative}.
We use AdamW optimizer~\cite{loshchilov2017decoupled} with a learning rate of 1e-5. We train our model for 40 epochs with a batch size of 32.
We use the RGB feature embedded by TSN~\cite{wang2016_TemporalSegmentNetworks} in this experiment. 
\\
\noindent
\textbf{Results.}
\begin{table}[t]
\begin{center}
\scalebox{1.0}{
\begin{tabular}{lcc}
        \toprule
        method & backbone & top-1 \\
        \cmidrule{1-3}
        RULSTM~\cite{furnari2019would}      & TSN & 13.1  \\
        Temporal Agg.~\cite{sener2020temporal} & TSN & 12.3  \\
        AVT~\cite{girdhar2021anticipative}    & TSN & 13.1 \\
        FUTR (ours) & TSN & 12.3  \\
        \bottomrule 
        \end{tabular}
        }
\vspace{-1mm}
\counterwithin{table}{section}
\renewcommand\thetable{S\arabic{table}}
\caption{\textbf{Performance comparison on EK55.} Although FUTR is designed for long-term action anticipation, the model is also effective in next action anticipation.}
\label{tab:EK55}
\end{center}
\vspace{-4mm}
\end{table}
The result is shown in Table~\ref{tab:EK55}. \ours obtains 12.3\%p at top-1 accuracy performing comparable with the state-of-the-art methods. We find that \ours is also effective for next action anticipation, although the model is designed for long-term action anticipation.


\section*{C. Additional Analysis}
\label{sec:C}
We conduct additional experiments for further analysis of the proposed method. In the following experiments, we evaluate our models on the Breakfast dataset with two observed ratios $\alpha \in \{0.2, 0.3\}$. Unless otherwise specified, all experimental settings are the same as those in Sec. 5.4.

\begin{table}[t]
\begin{center}
\renewcommand{\arraystretch}{0.8}
\scalebox{0.62}{
\begin{tabular}{wl{1.8cm}C{1cm}C{0.8cm}C{0.8cm}C{0.8cm}C{0.8cm}C{0.8cm}C{0.8cm}C{0.8cm}C{0.8cm}}
\toprule
\multirow{2}*[-0.5ex]{method} &\multirow{2}*[-0.5ex]{input} &\multicolumn{8}{c}{$\beta~(\alpha=0.2$)}\\
\cmidrule{3-10}
& &0.01 & 0.02 & 0.03 & 0.05& 0.1 & 0.2 & 0.3 & 0.5 \\
\cmidrule{1-10}
AVT~\cite{girdhar2021anticipative}&ViT&30.25 & 30.24 & 25.72 & 21.87& 14.22 & 10.69 & 8.49 & 5.83 \\
AVT~\cite{girdhar2021anticipative}&I3D& 26.13 & 22.03 & 20.24 & 13.52 & 17.84 & 13.20 & 9.01 & 4.61\\
\ccol\textbf{\ours~(ours)} &\ccol I3D &\ccol \textbf{51.16} & \ccol \textbf{44.34} &\ccol \textbf{40.84} &\ccol \textbf{40.56}   &\ccol \textbf{39.43} & \ccol \textbf{27.54} & \ccol \textbf{23.31} & \ccol \textbf{17.77} \\
\midrule
&&\multicolumn{8}{c}{$\beta~(\alpha=0.3$)}\\
\midrule
AVT~\cite{girdhar2021anticipative}&ViT & 30.93 & 30.62 & 27.85 & 23.60 & 18.28 & 13.51 & 9.65 & 7.35\\
AVT~\cite{girdhar2021anticipative}&I3D& 31.56 & 35.17 & 33.12 & 24.17 & 14.92 & 12.79 & 10.38 & 5.81\\
\ccol\textbf{\ours~(ours)} &\ccol I3D  &\ccol \textbf{42.20} & \ccol \textbf{38.67} & \ccol \textbf{38.56} & \ccol \textbf{36.44} &\ccol \textbf{35.15} & \ccol \textbf{24.86} &\ccol \textbf{24.22} &\ccol \textbf{15.26}\\
\bottomrule
\end{tabular}
}
\end{center}
\vspace{-4mm}
\counterwithin{table}{section}
\renewcommand\thetable{S\arabic{table}}
\caption{\textbf{Performance comparison with AVT on 50Salads.} FUTR outperforms AVT especially when predicting long-term action sequences.}
\vspace{-4mm}
\label{tab:avt}
\end{table}

\begin{figure}[t]
\vspace{3mm}
\centering
\includegraphics[width=\linewidth]{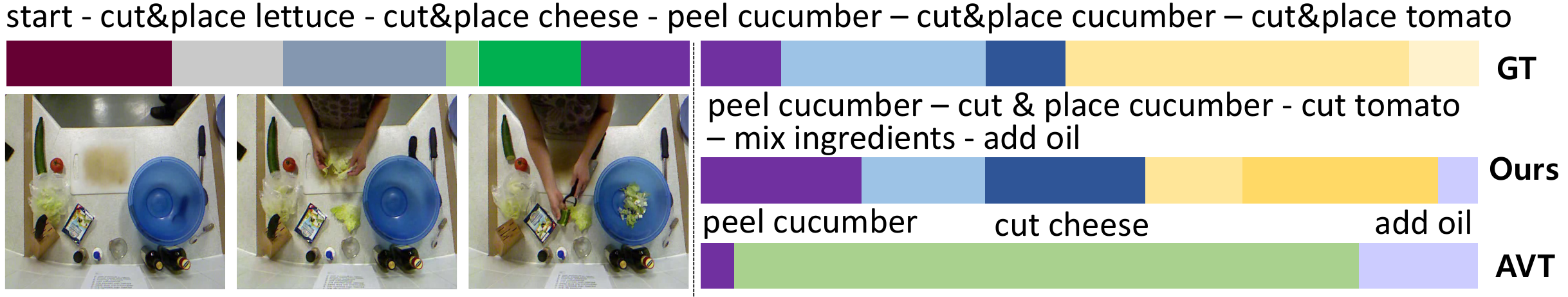}
\counterwithin{figure}{section}
\renewcommand\thefigure{S\arabic{figure}}
\vspace{-6mm}
\caption{\textbf{Qualitative results of FUTR vs. AVT on 50Salads.} Each color in the color bar indicates an action label written above. AVT becomes inaccurate in the prolong predictions while our method is consistently accurate.}
\label{fig:avt}
\end{figure}
\begin{figure}[t]
    \vspace{-1mm}
    \centering
        \begin{minipage}{\linewidth}
            \begin{subfigure}{\linewidth}
                \centering
                \scalebox{0.82}{
                \includegraphics[width=\linewidth]{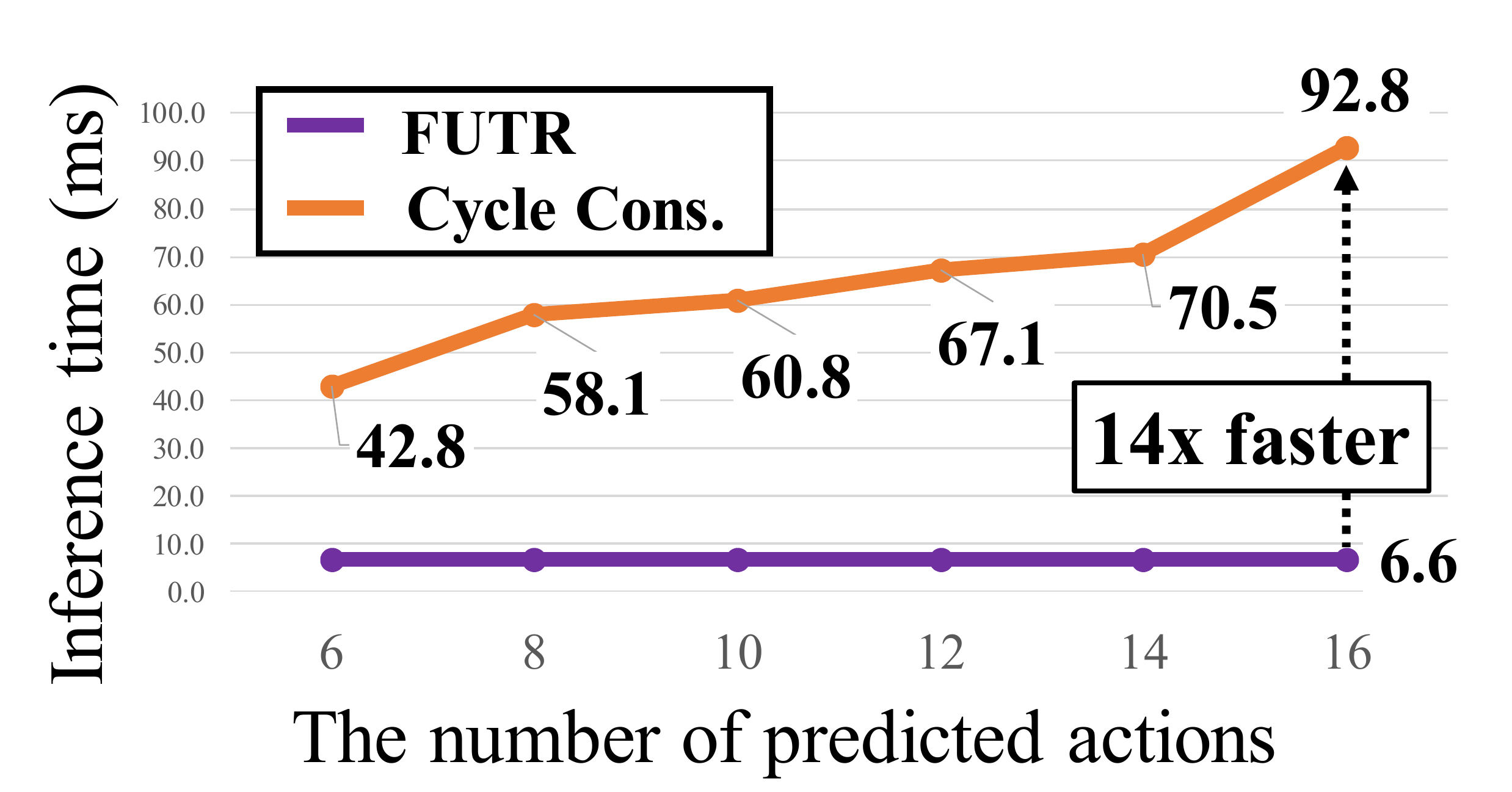}
                }
                \subcaption{\ours vs. Cycle Cons.~\cite{farha2020long}}
                \label{fig4:sub1}
            \end{subfigure}        
        \end{minipage}
        
        \vspace{2mm}
        
    \begin{minipage}{\linewidth}
            \begin{subfigure}{\linewidth}
                \centering
                \scalebox{0.8}{
                \includegraphics[width=\linewidth]{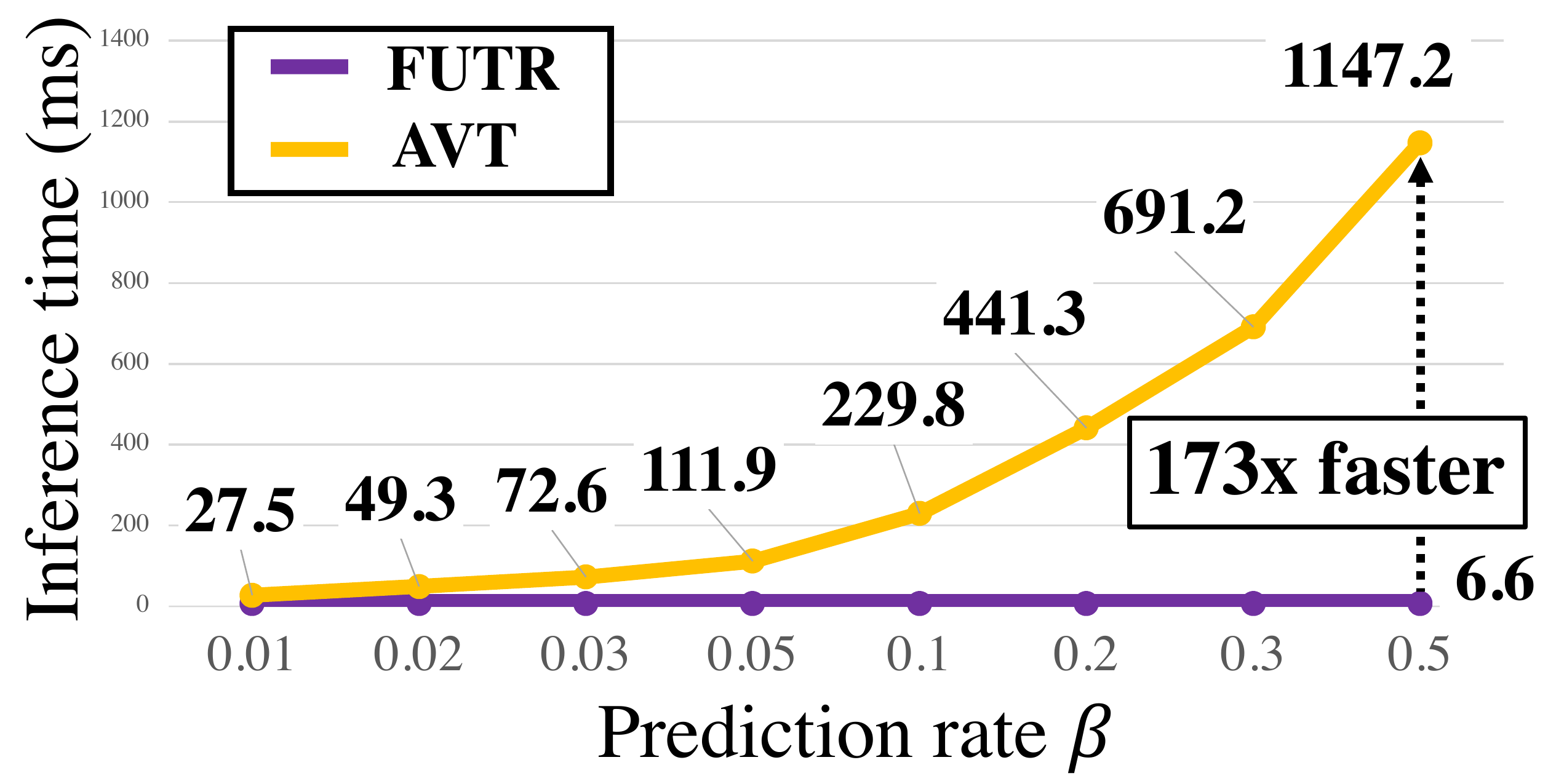}
                }
                \subcaption{\ours vs. AVT~\cite{girdhar2021anticipative}}
            \label{fig4:sub2}
            \end{subfigure}        
        \end{minipage}
    \vspace{-3.5mm}
    \counterwithin{figure}{section}
    \renewcommand\thefigure{S\arabic{figure}}
    \caption{\textbf{Inference time comparison with other methods}~\cite{farha2020long,girdhar2021anticipative}. Inference time of other methods linearly increases as the duration of the future action sequence to be predicted increases, \ie, the number of predicted actions or prediction rate $\beta$, increases, while that of FUTR is consistently fast.}
    \label{fig:inf}
    \vspace{-6mm}
\end{figure}

\begin{table*}[t]
\vspace{0mm}
\begin{center}
\scalebox{0.9}{
\begin{tabular}{C{1.8cm}C{1.1cm}C{1.1cm}C{1.1cm}C{1.1cm}C{1.1cm}C{1.1cm}C{1.1cm}C{1.1cm}}
            \toprule
            \multirow{2}*[-0.5ex]{$\gamma$} & \multicolumn{4}{c}{$\beta~(\alpha=0.2)$}&\multicolumn{4}{c}{$\beta~(\alpha=0.3)$} \\
            \cmidrule(r){2-5}
            \cmidrule(l){6-9}
            & 0.1 & 0.2 & 0.3 & 0.5& 0.1 & 0.2 & 0.3 & 0.5  \\
            \midrule
            0.25 & 24.36 & 21.66 &  20.62 & 20.10 & 30.31 &27.53 & 25.45 & 23.19 \\
            0.5 & 25.26 & 22.99 &  22.10 & 21.37 & 31.14 &28.25 & 25.91 & 23.85 \\
            \ccolg 1 & \ccol\textbf{27.70} & \ccol\textbf{24.55} & \ccol\textbf{22.83} & \ccol\textbf{22.04}&\ccolg \textbf{32.27} & \ccolg \textbf{29.88} &  \ccolg \textbf{27.49} & \ccolg \textbf{25.87}\\
            \bottomrule
            \end{tabular}}
\end{center}
\vspace{-4mm}
\counterwithin{table}{section}
\renewcommand\thetable{S\arabic{table}}
\caption{\textbf{Effectiveness of global cross-attention.} We set observed ratio from the recent past as $\gamma$ to show the effectiveness of exploiting global cross-attention at long-term action anticipation. }
\vspace{-1mm}
\label{tab:sup_globality_full}
\end{table*}


\begin{table*}[t!]
\vspace{1mm}
\begin{center}
\scalebox{0.9}{
\begin{tabular}{C{0.73cm}C{0.75cm}C{0.73cm}C{0.75cm}C{0.8cm}C{0.8cm}C{0.8cm}C{0.8cm}C{0.8cm}C{0.8cm}C{0.8cm}C{0.8cm}}
        \toprule
        \multicolumn{2}{c}{encoder}&\multicolumn{2}{c}{decoder} & \multicolumn{4}{c}{$\beta~(\alpha=0.2)$} & \multicolumn{4}{c}{$\beta~(\alpha=0.3)$}\\
        \cmidrule(r){1-2}
        \cmidrule{3-4}
        \cmidrule(l){5-8}
        \cmidrule(l){9-12}
        type&Loc.&type&Loc.& 0.1 & 0.2 & 0.3 & 0.5 & 0.1 & 0.2 & 0.3 & 0.5 \\
        \midrule
        -&-&learn&input& 21.79 & 20.14 & 19.88 & 18.25 & 26.53 &24.85 &25.04  &21.31 \\
        sine &input&learn&input& 21.29 & 19.55 & 19.11 & 18.23 & 27.40  &24.91 &24.13  &21.81\\
        learn &input&learn&input &23.79 & 21.37 & 20.49 & 19.62 & 30.80& 27.69 & 25.53 & 23.39\\
        \ccolg learn &\ccolg attn.&\ccolg learn&\ccolg attn.& \ccol\textbf{27.70} & \ccol\textbf{24.55} & \ccol\textbf{22.83} & \ccol\textbf{22.04} & \ccolg\textbf{32.27} &\ccolg\textbf{29.88} & \ccolg\textbf{27.49} & \ccolg \textbf{25.87}\\
        \bottomrule
        \end{tabular}
        }
\vspace{-1mm}
\counterwithin{table}{section}
\renewcommand\thetable{S\arabic{table}}
\caption{\textbf{Position embedding analysis.} Adding learnable positional embeddings before every attention layer performs the best.
}
\label{tab:pos_full}
\end{center}
\vspace{-4mm}
\end{table*}
\begin{table*}[t!]
\begin{center}
\scalebox{0.9}{
\begin{tabular}{C{1.1cm}C{1.1cm}C{1.1cm}C{0.8cm}C{0.8cm}C{0.8cm}C{0.8cm}C{0.8cm}C{0.8cm}C{0.8cm}C{0.8cm}}
            \toprule
            \multicolumn{3}{c}{model} & \multicolumn{4}{c}{$\beta~(\alpha=0.2)$}&\multicolumn{4}{c}{$\beta~(\alpha=0.3)$} \\ 
            \cmidrule(r){1-3}
            \cmidrule{4-7}
            \cmidrule(l){8-11}
            $L^\mathrm{E}$&$L^\mathrm{D}$&$D$ & 0.1 & 0.2 & 0.3 & 0.5&0.1 & 0.2 & 0.3 & 0.5 \\
            \midrule
            1&1& 128 &24.02&21.00& 19.71 & 19.39 & 29.38 &26.51 & 25.06 & 23.83 \\
            \ccolg 2&\ccolg 1& \ccolg 128 & \ccol\textbf{27.70} & \ccol\textbf{24.55} & \ccol22.83 &\ccol\textbf{22.04}& \ccolg 32.27 & \ccolg \textbf{29.88} &  \ccolg27.49 & \ccolg \textbf{25.87}\\
            3&1& 128 & 24.78 & 22.78 & 21.46 & 20.53 & 30.44 & 27.61 & 25.73 & 23.75\\
            3&2& 128 & 26.72 & 23.82 & 22.57 & 21.29 & 32.55 & 29.20 & 26.59 & 24.92\\
            3&3& 128 & 26.68 & 23.41 & 22.14 & 21.56 & \textbf{33.06} & 29.14 & \textbf{28.12} & 24.93\\
            4&4& 128 & 26.77 & 23.60 & 22.92 & 21.24 & 31.35 & 28.58 & 27.04 & 24.73\\
            5&5& 128 & 26.75 & 24.23 & \textbf{23.55} & 21.16 & 32.68 & 29.38 & 28.05 & 24.89\\
            \midrule
            2&1& 64 & 24.78 & 21.81 & 20.56 & 19.88 & 29.92 & 27.03& 26.29 & 23.53\\
            2&1& 256 & 24.56 & 21.62 & 21.35 & 19.41 & 31.19 & 26.03 & 26.07 & 24.29\\
            2&1& 512 & 19.82 & 17.50 & 18.07 & 16.31 & 23.68 & 22.18 & 23.57 & 22.56\\
            \bottomrule
            \end{tabular}}
\end{center}
\vspace{-5mm}
\counterwithin{table}{section}
\renewcommand\thetable{S\arabic{table}}
\caption{\textbf{Model analysis.} We study the number of encoder layers $L^\textrm{E}$, the number of decoder layers $L^\textrm{D}$, and hidden dimension $D$ of our model. We show the robustness of our methods over the number of layers, and find that the optimal hidden dimension $D$ is 128. }
\vspace{0mm}
\label{tab:model_full}
\end{table*}

\begin{table*}[t!]
\begin{center}
\scalebox{0.9}{
\begin{tabular}{C{1.7cm}C{1.1cm}C{1.1cm}C{1.1cm}C{1.1cm}C{1.1cm}C{1.1cm}C{1.1cm}C{1.1cm}}
            \toprule
            \multirow{2}*[-0.5ex]{loss} & \multicolumn{4}{c}{$\beta~(\alpha=0.2)$}&\multicolumn{4}{c}{$\beta~(\alpha=0.3)$} \\ 
            \cmidrule(r){2-5}
            \cmidrule(l){6-9}
            & 0.1 & 0.2 & 0.3 & 0.5& 0.1 & 0.2 & 0.3 & 0.5  \\
            \midrule
            L1 & 23.90 & 21.46 &  20.76 & 20.08 & 30.72 & 27.28 & 26.08 & 23.90\\
            smooth L1 &23.07 & 23.36 &  \textbf{22.88} & 20.74& 29.96 & 27.20 & 24.78 & 23.18\\
            \ccolg L2 & \ccol\textbf{27.70} & \ccol\textbf{24.55} & \ccol22.83 & \ccol\textbf{22.04}&\ccolg \textbf{32.27} & \ccolg \textbf{29.88} &  \ccolg \textbf{27.49} & \ccolg \textbf{25.87}\\
            \bottomrule
            \end{tabular}}
\end{center}
\vspace{-4mm}
\counterwithin{table}{section}
\renewcommand\thetable{S\arabic{table}}
\caption{\textbf{Duration loss analysis.} We find that utilizing L2 loss as our duration loss $\mathcal{L}^\mathrm{duration}$ shows better performance over L1 loss and Smooth L1 loss.}
\vspace{-4mm}
\label{tab:lenloss_full}
\end{table*}

\noindent
\textbf{Comparison with AVT.}
The core difference between AVT~\cite{girdhar2021anticipative} and \ours lies in the transformer architecture and the parallel decoding. 
While AVT uses a simple decoder that predicts the next action within a few seconds considering only the previous actions via masked self-attention, \ours adopts a full-fledged decoder that predicts the whole sequence of actions in parallel by examining long-term relations of the actions via self-attention and cross-attention. 
AVT is also capable of anticipating long-term actions by unrolling the decoder iteratively, but it remains the drawbacks of error accumulation and slow inference speed. 
To validate our claim, we compare our method with AVT\footnote{We evaluate two AVT models trained on 50 Salads according to different types of backbone networks: AVT with ViT, where the trained model is available on their official website (\url{www.github.com/facebookresearch/AVT}), and AVT with I3D, where the model is trained by using their official codes.}
on long-term action anticipation.

Table~\ref{tab:avt} shows the results of long-term action anticipation of both models. 
Since AVT is built for next action anticipation, we also adjust the prediction rate $\beta$ ranging from 0.01 to 0.5. 
As $\beta$ becomes smaller, the prediction results are closely related to next action anticipation.
We find that AVT performs inferior to \ours, especially when predicting long-term sequences. AVT is accurate for the early frames but becomes inaccurate in the prolonged predictions as shown in Fig.~\ref{fig:avt}.
\\
\noindent
\textbf{Inference time comparison.}
We compare inference time of FUTR to that of the Cycle Cons.~\cite{farha2020long} and AVT~\cite{girdhar2021anticipative} in Fig.~\ref{fig:inf}.
The vertical axis indicates the inference time~(ms) and the horizontal axis indicates the number of predicted actions for Cycle Cons. and the prediction rate $\beta$ for AVT.
The inference time of FUTR is consistently fast while that of Cycle Cons. and AVT linearly increases as the duration of the predicted sequence increases.
From this experiment, we find that \ours is 14$\times$ faster than Cycle Cons. when predicting 16 actions and 173$\times$ faster than AVT when $\beta$ is set to 0.5.
The results show the efficiency of the parallel decoding for long-term action anticipation.



\begin{table*}[hbt!]
\vspace{2mm}
\begin{center}
\scalebox{0.9}{
\begin{tabular}{C{1.4cm}C{0.8cm}C{1cm}C{1cm}C{1cm}C{1cm}C{1cm}C{1cm}C{1cm}C{1cm}C{1cm}}
\toprule
\multirow{2}*[-0.5ex]{method} &\multirow{2}*[-0.5ex]{AR} &\multirow{2}*[-0.5ex]{\shortstack{causal \\ mask}} &\multicolumn{4}{c}{$\beta~(\alpha=0.2)$} & \multicolumn{4}{c}{$\beta~(\alpha=0.3)$}\\
\cmidrule(r){4-7}
\cmidrule(l){8-11}
&&& 0.1 & 0.2 & 0.3 & 0.5 & 0.1 & 0.2 & 0.3 & 0.5\\
\midrule
FUTR-A&\checkmark&\checkmark & 20.31 & 18.37 &  17.69 & 16.31 & 25.43 & 24.02 & 23.43 & 21.08 \\
FUTR-M&-& \checkmark& 25.27 & 22.41 & 21.39 & 20.86 &  31.82 & 28.55 & 26.57 &  24.17 \\
\ccol FUTR&\ccol-&\ccol- & \ccol\textbf{27.70} & \ccol\textbf{24.55} & \ccol\textbf{22.83} & \ccol\textbf{22.04} & \ccol\textbf{32.27} &\ccol\textbf{29.88} & \ccol\textbf{27.49} &  \ccol\textbf{25.87}\\
\bottomrule
\end{tabular}
}
\end{center}
\vspace{-4mm}
\counterwithin{table}{section}
\renewcommand\thetable{S\arabic{table}}
\caption{\textbf{Parallel decoding vs. autoregressive decoding.} }
\label{tab:parallel_full}
\end{table*}

\begin{table*}[hbt!]
\vspace{2mm}
\begin{center}
\scalebox{0.9}{
\begin{tabular}{C{1.4cm}C{1.4cm}C{1.1cm}C{1.1cm}C{1.1cm}C{1.1cm}C{1.1cm}C{1.1cm}C{1.1cm}C{1.1cm}}
\toprule
\multirow{2}*[-0.5ex]{encoder} &\multirow{2}*[-0.5ex]{decoder} &\multicolumn{4}{c}{$\beta~(\alpha=0.2)$} & \multicolumn{4}{c}{$\beta~ (\alpha=0.3)$} \\
\cmidrule( r){3-6}
\cmidrule( l){7-10}
&& 0.1 & 0.2 & 0.3 & 0.5 & 0.1 & 0.2 & 0.3 & 0.5 \\
\cmidrule{1-10}
LSA&LSA & 21.97 & 19.20 &  18.04 & 18.19 & 27.70 & 24.39 & 23.18 &  21.60\\
GSA&LSA& 25.25 & 22.88 & 21.09 & 19.73 & 30.15 & 27.51 & 25.62 &  23.28\\
LSA&GSA& 22.99 & 20.39 & 19.15 & 18.60 & 28.37 & 25.08 & 24.03 &  22.28\\
\ccol GSA& \ccol GSA & \ccol\textbf{27.70} & \ccol\textbf{24.55} & \ccol\textbf{22.83} & \ccol\textbf{22.04} & \ccol\textbf{32.27} &\ccol\textbf{29.88} & \ccol\textbf{27.49} &  \ccol\textbf{25.87}\\
\bottomrule
\end{tabular}
}
\end{center}
\vspace{-4mm}
\counterwithin{table}{section}
\renewcommand\thetable{S\arabic{table}}
\caption{\textbf{Global self-attention~(GSA) vs. local self-attention~(LSA).} }
\label{tab:LSAity_full}
\end{table*}


\begin{table*}[hbt!]
\vspace{2mm}
\begin{center}
\scalebox{0.9}{
\begin{tabular}{C{1.5cm}C{1.7cm}C{1.6cm}C{0.8cm}C{0.8cm}C{0.8cm}C{0.8cm}C{0.8cm}C{0.8cm}C{0.8cm}C{0.8cm}}
\toprule
\multirow{2}*[-0.5ex]{\shortstack{method}} &\multirow{2}*[-0.5ex]{\shortstack{GT Assign.}} &\multirow{2}*[-0.5ex]{regression} &\multicolumn{4}{c}{$\beta~(\alpha=0.2)$} & \multicolumn{4}{c}{$\beta~(\alpha=0.3)$} \\
\cmidrule( r){4-7}
\cmidrule( l){8-11}
&&& 0.1 & 0.2 & 0.3 & 0.5 & 0.1 & 0.2 & 0.3 & 0.5 \\
\cmidrule{1-11} 
FUTR-S & sequential & start-end & 23.87 & 19.86 & 18.58 & 18.05 & 29.15 & 25.51 & 24.20 & 21.43\\
FUTR-H & Hungarian & start-end & 22.05 & 20.18 &  18.63 & 17.31 & 25.26 & 23.85 & 22.63 & 21.45\\
\ccol FUTR & \ccol sequential & \ccol duration & \ccol\textbf{27.70} & \ccol\textbf{24.55} & \ccol\textbf{22.83} & \ccol\textbf{22.04} & \ccol\textbf{32.27} &\ccol\textbf{29.88} & \ccol\textbf{27.49} &  \ccol\textbf{25.87}\\
\bottomrule
\end{tabular}
}
\end{center}
\vspace{-4mm}
\counterwithin{table}{section}
\renewcommand\thetable{S\arabic{table}}
\caption{\textbf{Output structuring.} }
\label{tab:setpred_full}
\end{table*}




\begin{table*}[hbt!]
\vspace{2mm}
\begin{center}
\scalebox{0.9}{
\begin{tabular}{C{0.9cm}C{1.05cm}C{1.4cm}C{1cm}C{1cm}C{1cm}C{1cm}C{1cm}C{1cm}C{1cm}C{1cm}}
\toprule
\multicolumn{3}{c}{loss} &\multicolumn{4}{c}{$\beta~(\alpha=0.2)$} & \multicolumn{4}{c}{$\beta~(\alpha=0.3)$} \\ 
\cmidrule( r){1-3}
\cmidrule{4-7}
\cmidrule( l){8-11}
$\mathcal{L}^{\mathrm{seg}}$&$\mathcal{L}^{\mathrm{action}}$&$\mathcal{L}^{\mathrm{duration}}$ & 0.1 & 0.2 & 0.3 & 0.5 & 0.1 & 0.2 & 0.3 & 0.5 \\
\cmidrule( r){1-11}
-&\checkmark&\checkmark & 25.60 & 22.13 &  21.95 & 20.86 & 28.31 & 25.85 & 24.91 & 22.50\\
\ccol\checkmark&\ccol\checkmark&\ccol\checkmark & \ccol\textbf{27.70} & \ccol\textbf{24.55} & \ccol\textbf{22.83} &\ccol\textbf{22.04} &  \ccol\textbf{32.27} &  \ccol\textbf{29.88} &  \ccol\textbf{27.49} &  \ccol\textbf{25.87}\\
\bottomrule
\end{tabular}
}
\end{center}
\vspace{-4mm}
\counterwithin{table}{section}
\renewcommand\thetable{S\arabic{table}}
\caption{\textbf{Loss ablations.} }
\label{tab:loss_full}
\end{table*}

\begin{table*}[hbt!]
\begin{center}
\scalebox{0.9}{
\begin{tabular}{C{2.4cm}C{1.2cm}C{1.2cm}C{1.2cm}C{1.2cm}C{1.2cm}C{1.2cm}C{1.2cm}C{1.2cm}}
        \toprule
        \multirow{2}*[-0.5ex]{$M$}&\multicolumn{4}{c}{$\beta~(\alpha=0.2)$}& \multicolumn{4}{c}{$\beta~(\alpha=0.3)$} \\ 
        \cmidrule(r){2-5}
        \cmidrule(l){6-9}
        & 0.1 & 0.2 & 0.3 & 0.5 &0.1 & 0.2 & 0.3 & 0.5 \\
        \midrule
        6 & 24.63 & 21.74 & 20.99 & 19.67 & 29.95 & 26.47 & 25.46 & 23.27 \\
        7 & 24.40 & 22.13 & 21.59 &20.28 &30.03 & 27.94& 27.00 & 24.23 \\
        \ccolg 8 &\ccol \textbf{27.70} & \ccol \textbf{24.55} & \ccol \textbf{22.83} & \ccol \textbf{22.04} & \ccolg \textbf{32.27} & \ccolg \textbf{29.88} &  \ccolg 27.49 & \ccolg \textbf{25.87}\\
        9 & 24.21 & 22.47 & 21.56 & 20.94 & 31.24 & 28.65& 26.87 & 24.95 \\
        10 & 24.61 & 21.79 & 20.90 & 19.91 &31.32 & 28.86& \textbf{27.74} & 25.01 \\
        \bottomrule
        \end{tabular}}
\end{center}
\vspace{-6mm}
\counterwithin{table}{section}
\renewcommand\thetable{S\arabic{table}}
\caption{\textbf{Number of action queries.}}
\label{tab:query_full}
\end{table*}

\noindent
\textbf{Effectiveness of global cross-attention.} 
To evaluate the importance of modeling long-term dependencies between the observed frames and the action queries during the decoding stage, we measure the performance by gradually increasing the number of cross-attended frames from the most recent frame to the farthest one.
For notational simplicity, we establish the ratio of the cross-attended frames $\gamma$ ranging from 0.25 to 1, adjusting the number of observed frames starting from the recent past; the cross-attention layer in the decoder only attends to the most recent $\gamma T^\textrm{O}$ frames during the decoding stage.
Note that $\gamma = 1$, our default setting, indicates that the decoder attends to the whole video frames to anticipate actions.

Table~\ref{tab:sup_globality_full} summarizes the results of the effect of global attention in the cross-attention layers. As we gradually increase the $\gamma$ from 0.25 to 1, the overall accuracy significantly increases by 2.0-3.3\%p.
This demonstrates the efficacy of modeling global interactions between the observed frames in the past and the action queries in the future for long-term action anticipation.
\\
\noindent
\textbf{Position embedding analysis.}
In Table~\ref{tab:pos_full}, we investigate various combinations of different types and locations of the positional embeddings.
From the 1$^\mathrm{st}$ to the 3$^\mathrm{rd}$ rows, we compare three types of position embeddings in the encoder layers: none, sinusoidal, and learnable position embeddings. Here, we fix the position embedding of the decoder as learnable embedding, which is added before going into the attention layers. We find that using learnable position embeddings in the encoder is effective.
Then we change the location of the position embeddings to be learned in the attention layers, obtaining additional accuracy improvements. In this experiment, we find that position embedding learned at the attention layer is effective for our model.
\\
\noindent
\textbf{Model analysis.}
Table~\ref{tab:model_full} summarizes the results of the model ablations, according to the number of encoder layers $L^\textrm{E}$, the number of decoder layers $L^\textrm{D}$, and hidden dimension $D$.
We find that the performance is saturated when we use more than two encoder layers and one decoder layer. Thus we set $L^\textrm{E}=2$ and $L^\textrm{D}=1$ as our default number of encoder layers and decoder layers, respectively.
We also evaluate our model by varying the channel dimension $D$ and find that setting $D$ to 128 performs the best; too small $D$ restricts the representation power of the model while too large $D$ causes overfitting problems.

\noindent
\textbf{Duration loss analysis.}
In Table~\ref{tab:lenloss_full}, we evaluate our duration loss $\mathcal{L}^\mathrm{duration}$ of Eq.~(14). Instead of L2 loss, we use L1 loss and Smooth L1 loss~\cite{girshick2015fast} in this experiment. The results show that applying L2 loss shows better performance over the L1 loss and smooth L1 loss. 
Since L2 loss is more robust to outliers than L1 loss and smooth L1 loss, we find that applying L2 loss is effective in the proposed method.


\section*{D. Additional Results}
\label{sec:D}
In Tables~\ref{tab:parallel_full}-\ref{tab:query_full}, we provide the overall experimental results in Sec.~5.4 with two observation ratios $\alpha \in \{ 0.2, 0.3\}$.
We find that overall experimental tendencies with the two observation ratios are similar although the experimental setup with $\alpha = 0.2$ is more challenging.


\section*{E. Qualitative Results}
\label{sec:E}
\begin{figure*}[t]
    \centering
    \counterwithin{figure}{section}
    \renewcommand\thefigure{C\arabic{figure}}
    \begin{subfigure}[t]{0.69\linewidth}
    \includegraphics[width=\columnwidth]{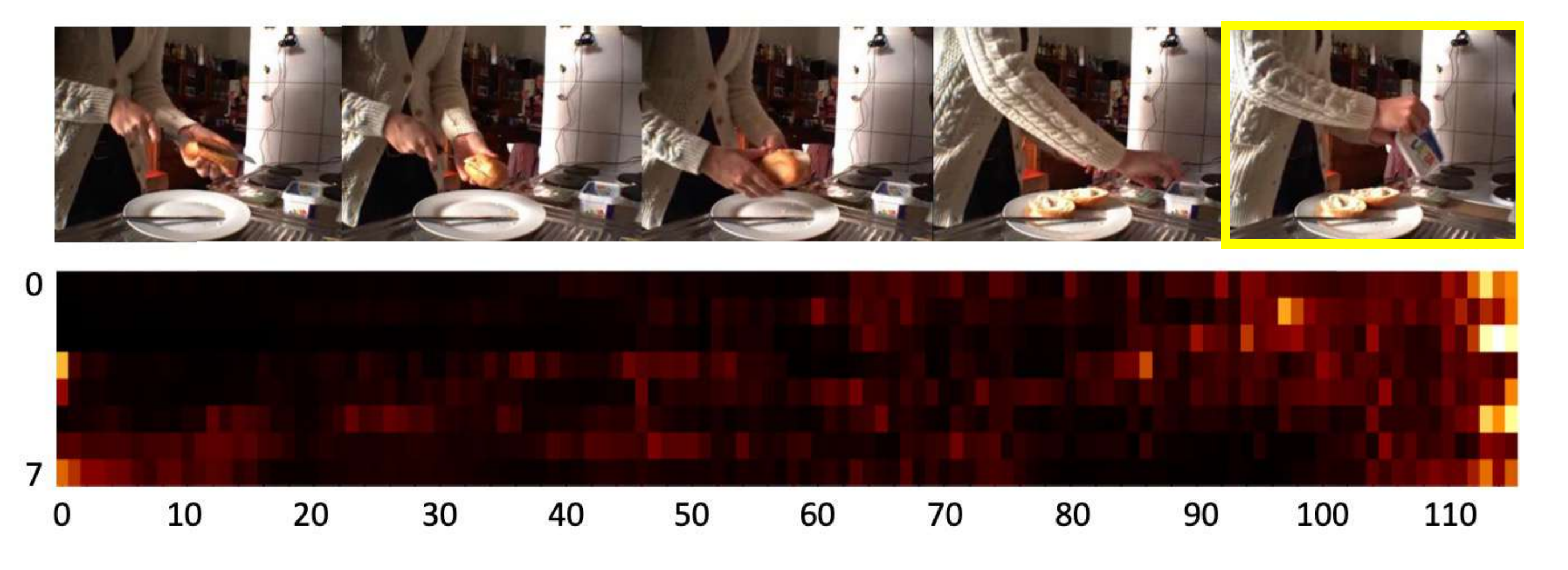}
    \counterwithin{figure}{section}
    \renewcommand\thefigure{S\arabic{figure}}
    \vspace{-5mm}
    \caption{Activity: Sandwich}
    \label{fig:s4a}
    \end{subfigure}
    
    \vspace{3mm}
    
    \begin{subfigure}[t]{0.75\linewidth}
    \includegraphics[width=\columnwidth]{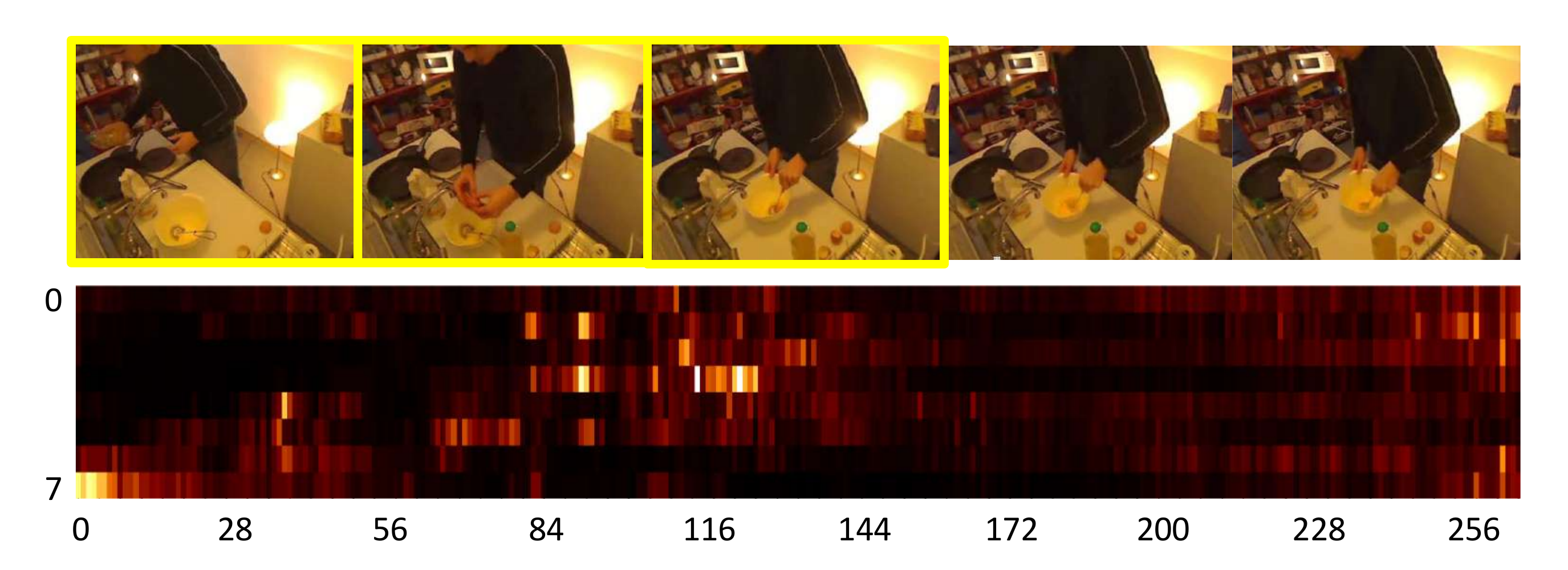}
    \counterwithin{figure}{section}
    \renewcommand\thefigure{S\arabic{figure}}
    \vspace{-6mm}
    \caption{Activity: Scramble egg}
    \label{fig:s4b}
    \end{subfigure}
    
    \vspace{3mm}
    
    \begin{subfigure}[t]{0.75\linewidth}
    \includegraphics[width=\columnwidth]{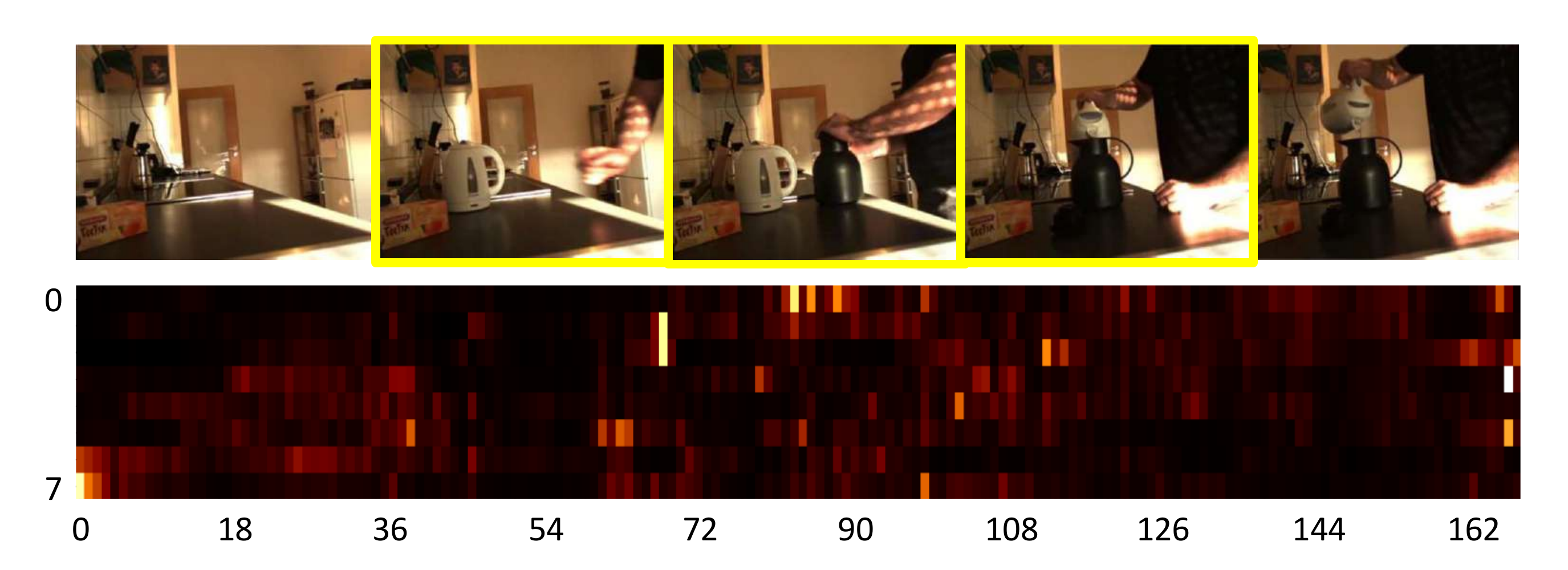}
    \counterwithin{figure}{section}
    \renewcommand\thefigure{S\arabic{figure}}
    \vspace{-6mm}
    \caption{Activity: Tea}
    \label{fig:s4d}
    \end{subfigure}
    
    \vspace{3mm}
    
    \begin{subfigure}[t]{0.75\linewidth}
    \includegraphics[width=\columnwidth]{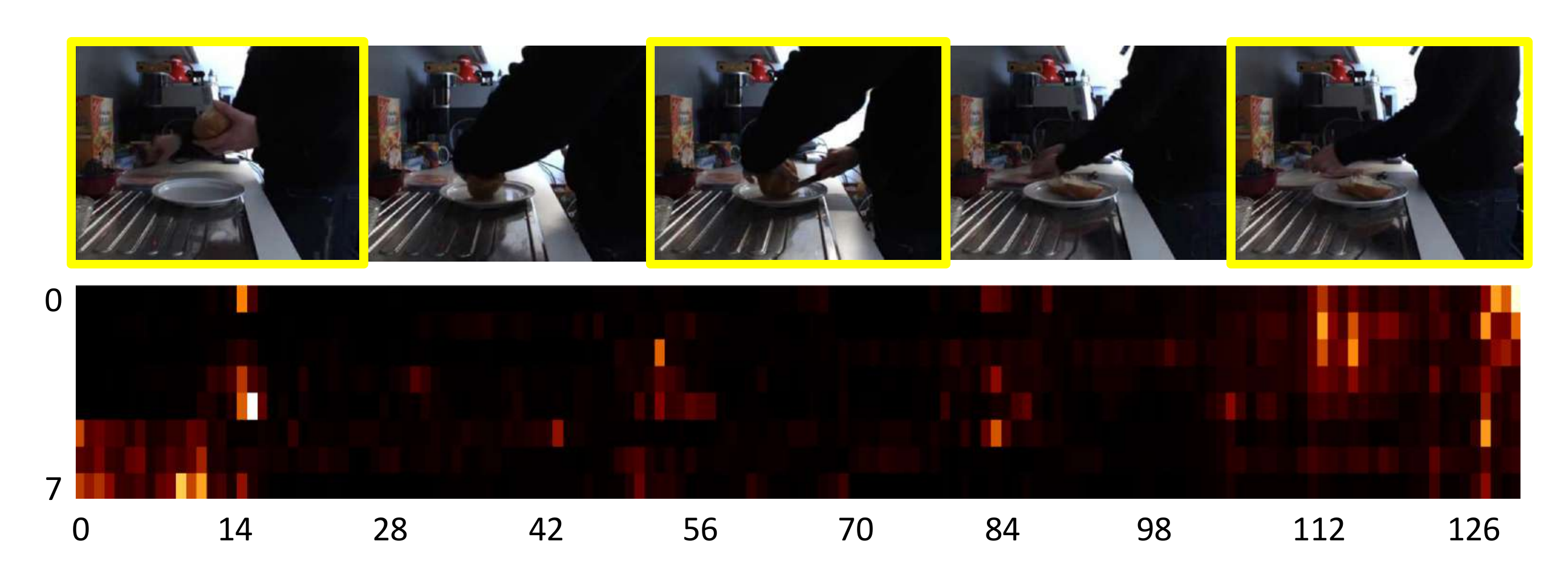}
    \counterwithin{figure}{section}
    \renewcommand\thefigure{S\arabic{figure}}
    \vspace{-6mm}
    \caption{Activity: Sandwich}
    \label{fig:s4e}
    \end{subfigure}
\counterwithin{figure}{section}
\renewcommand\thefigure{S\arabic{figure}}
\caption{\textbf{Cross-attention map visualization on Breakfast}. The vertical and horizontal axis indicates the decoder queries and observed frames, respectively. The brighter color indicates a higher attention score. RGB frames above the attention map are sampled uniformly from the video.
We emphasize the frames with high attention scores with yellow box and other frames are uniformly sampled. Best viewed in color.}
\vspace{-2mm}
 \label{fig:sup_attn}
\end{figure*}

We plot additional visualization results of the cross attention map of the decoder in Fig.~\ref{fig:sup_attn}.
Each subfigure contains sampled frames from videos and attention map visualizations below.
We also highlight the frames with the yellow box where corresponding attention scores are highly activated.
From this experiment, we find that action query in our method attends dynamically to the input visual features, which utilize fine-grained visual features from the entire past visual features.

\begin{figure*}[t]
    \centering
    \counterwithin{figure}{section}
    \renewcommand\thefigure{C\arabic{figure}}
    \begin{subfigure}[t]{0.95\linewidth}
    \includegraphics[width=\columnwidth]{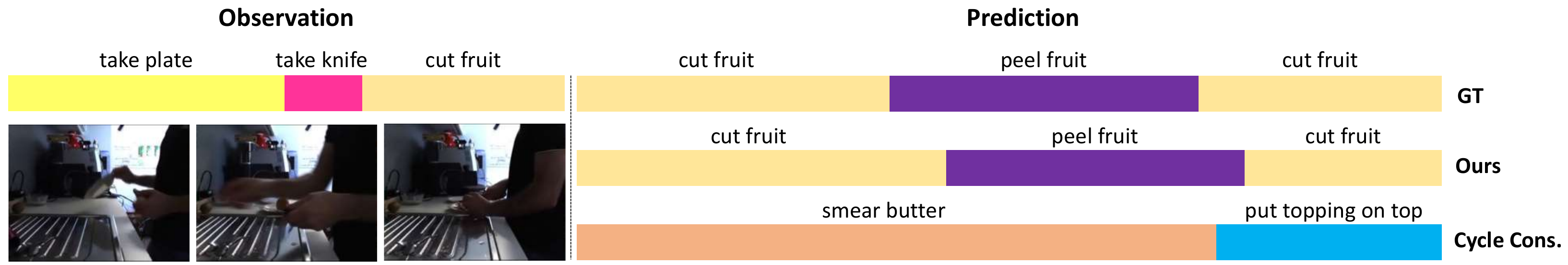}
    \counterwithin{figure}{section}
    \renewcommand\thefigure{S\arabic{figure}}
    \vspace{-4mm}
    \caption{Activity: salad}
    \label{fig:s5a}
    \end{subfigure}
    
    \begin{subfigure}[t]{0.95\linewidth}
    \includegraphics[width=\columnwidth]{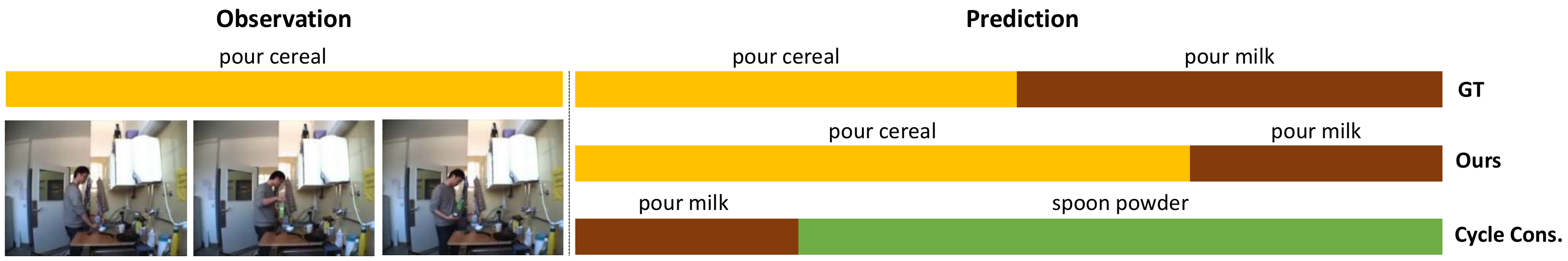}
    \counterwithin{figure}{section}
    \renewcommand\thefigure{S\arabic{figure}}
    \vspace{-4mm}
    \caption{Activity: cereal}
    \label{fig:s5b}
    \end{subfigure}

    \begin{subfigure}[t]{0.95\linewidth}
    \includegraphics[width=\columnwidth]{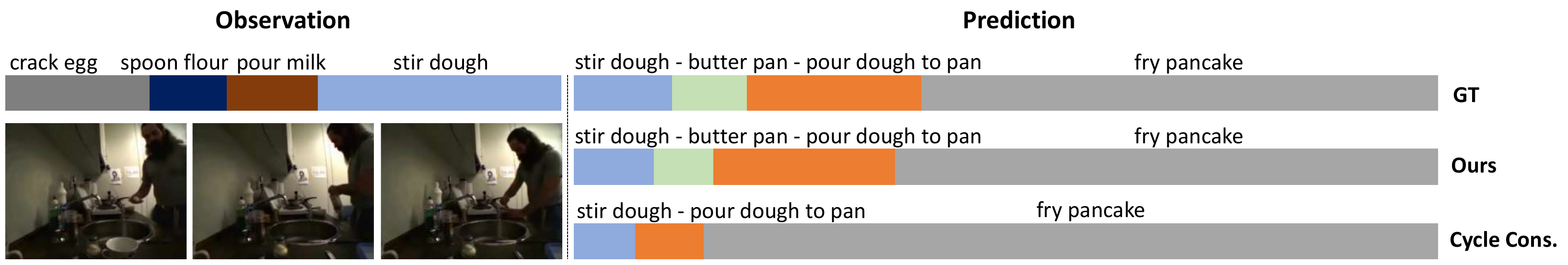}
    \counterwithin{figure}{section}
    \renewcommand\thefigure{S\arabic{figure}}
    \vspace{-4mm}
    \caption{Activity: pancake}
    \label{fig:s5c}
    \end{subfigure}
    
    \begin{subfigure}[t]{0.95\linewidth}
    \includegraphics[width=\columnwidth]{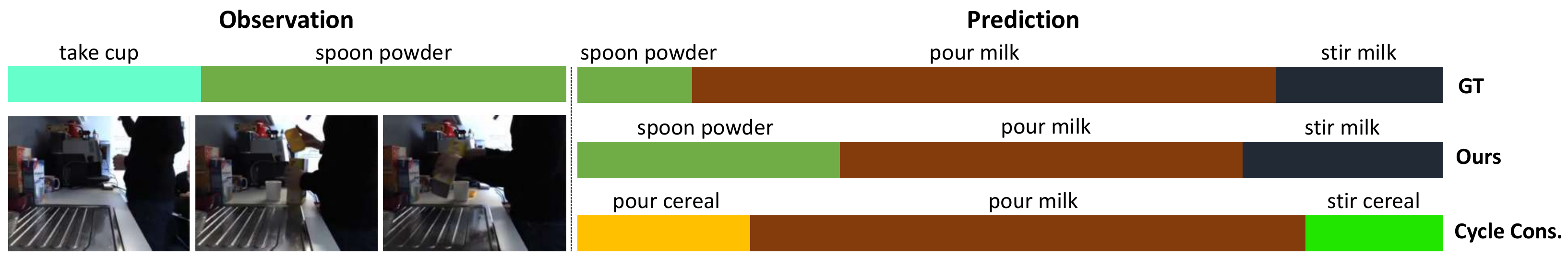}
    \counterwithin{figure}{section}
    \renewcommand\thefigure{S\arabic{figure}}
    \vspace{-4mm}
    \caption{Activity: milk}
    \label{fig:s5d}
    \end{subfigure}
    
    \begin{subfigure}[t]{0.95\linewidth}
    \includegraphics[width=\columnwidth]{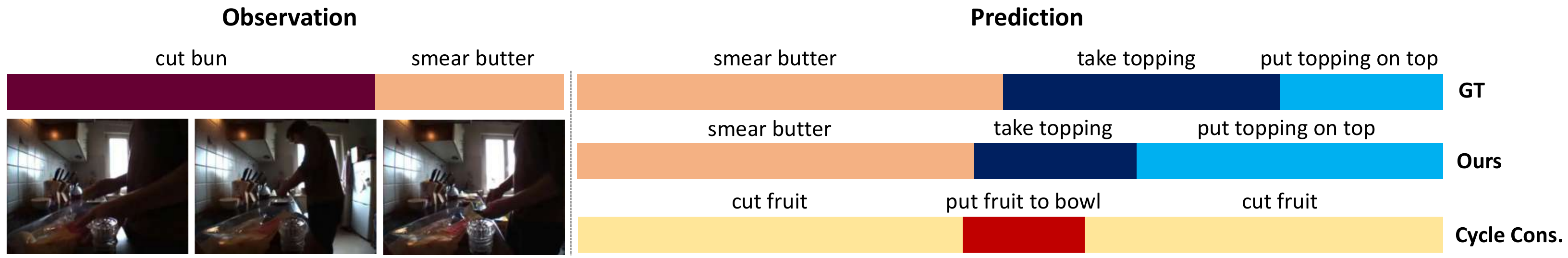}
    \counterwithin{figure}{section}
    \renewcommand\thefigure{S\arabic{figure}}
    \vspace{-4mm}
    \caption{Activity: pancake}
    \label{fig:s5e}
    \end{subfigure}
\vspace{-2mm}
\counterwithin{figure}{section}
\renewcommand\thefigure{S\arabic{figure}}
\caption{\textbf{Qualitative results on Breakfast}. Each subfigure visualizes the ground truth label and the prediction results of the \ours and cycle consistency model proposed from Farha~\etal~\cite{farha2020long}. We set $ \alpha $ as 0.3 and $ \beta $ as 0.5 in this experiment. We decode action labels and durations to the frame-wise action classes. Each color in the color bar indicates an action label written above. Best viewed in color.}
\vspace{40mm}
 \label{fig:sup_qual}
\end{figure*}

We also provide more qualitative results of our predictions over cycle consistency model~\cite{farha2020long} in Fig.~\ref{fig:sup_qual}.


\vspace{-1mm}
\section*{F. Discussion}
We have proposed an end-to-end attention network for long-term action anticipation, which effectively leverages global interactions in videos enabling accurate and fast inference for long-term action anticipation. We have demonstrated the effectiveness of the \ours through extensive experiments, but there exists much room for improvement. 
\\
\textbf{Limitations.}
First, the efficiency of \ours could be further improved. For example, linear attention mechanisms~\cite{katharopoulos2020transformers, wang2020linformer, choromanski2020rethinking} or sparse attention mechanisms~\cite{beltagy2020longformer,zaheer2020big} could reduce both computation and memory complexity of \ours, enabling efficient long-term video understanding.
Second, considering that our encoder is a separate action segmentation network, the proposed architecture is a unified network that can handle both long-term action anticipation and action segmentation task at once.
Although we focus on long-term action anticipation in this paper, we can integrate our models with other action segmentation methods~\cite{lea2017temporal,wang2020boundary,ishikawa2021alleviating,farha2019ms,yi2021asformer} to solve both action segmentation and long-term action anticipation task altogether in the same framework.
We leave this as our future work.
\\
\textbf{Societal impact.} Since our model is proposed to anticipate future actions and durations by observing past videos, our model can be used for predicting potential actions from people and can be applied to the surveillance system.

\clearpage
\clearpage
\vspace{40mm}

\end{document}